\documentclass[manuscript,screen,nonacm]{acmart}
\usepackage{amsmath,amsfonts}

\usepackage{algorithmic}
\usepackage{diagbox}
\usepackage{url}
\usepackage{caption}
\usepackage{subcaption}
\usepackage{booktabs}
\usepackage{hyperref}
\usepackage{color}
\usepackage{xcolor}
\usepackage{colortbl}
\usepackage[flushleft]{threeparttable}
\usepackage{tikz}
\usepackage{xspace}
\usepackage{bbding} 
\usepackage{romannum}
\usepackage{enumitem}
\usepackage[ruled,linesnumbered]{algorithm2e}

\SetKwInOut{Parameter}{Hyper-parameter}
\SetKwIF{If}{ElseIf}{Else}{if}{}{else if}{else}{end if}%
\SetKwFor{While}{while}{}{end while}%
\SetKwRepeat{Do}{do}{while}
\SetKwInOut{Parameter}{parameter}
\SetKwIF{If}{ElseIf}{Else}{if}{}{else if}{else}{end if}%
\SetKwFor{While}{while}{}{end while}%
\SetKwRepeat{Do}{do}{while}
\SetKwInOut{Parameter}{Hyper-parameter}

\newcommand{\method}{LAD\xspace}
\newcommand{\newmethod}{MultiLAD\xspace}
\newcommand{\modtextbf}[1]{\noindent{\textbf{#1:}~}}

\newcommand{\etc}{et al.}
\renewcommand{\vec}[1]{\ensuremath{\mathbf{#1}}}

\newcommand{\mat}[1]{\ensuremath{\mathbf{#1}}}

\newcommand{\ten}[1]{\mat{\ensuremath{\boldsymbol{\mathcal{#1}}}}}

\newcommand{\changed}[1]{\textcolor{black}{#1}}

\AtBeginDocument{%
  }

\setcopyright{acmcopyright}
\copyrightyear{2022}
\acmYear{2022}
\acmDOI{XXXXXXX.XXXXXXX}

\acmPrice{15.00}
\acmISBN{978-1-4503-XXXX-X/18/06}

\begin{document}

\pagenumbering{arabic}

\title[Laplacian Change Point Detection for Multi-view Dynamic Graphs]{Laplacian Change Point Detection for Single and Multi-view Dynamic Graphs}



\author{Shenyang~Huang}
\email{shenyang.huang@mail.mcgill.ca}
\affiliation{
  \institution{Mila,~McGill University}
  \country{Canada}
}
\author{Samy~Coulombe}
\affiliation{
  \institution{McGill University}
  \country{Canada}
}
\author{Yasmeen~Hitti}
\affiliation{
  \institution{Mila,~McGill University}
  \country{Canada}
}
\author{Reihaneh~Rabbany}
\affiliation{
  \institution{Mila, McGill University - CIFAR AI chair}
  \country{Canada}
}
\author{Guillaume~Rabusseau}

\affiliation{
   \institution{ Mila \& DIRO, Université de Montréal - CIFAR AI chair}
   \country{Canada}
   }

\renewcommand{\shortauthors}{Huang et al.}

\begin{abstract}
Dynamic graphs are rich data structures that are used to model complex relationships between entities over time. In particular, anomaly detection in temporal graphs is crucial for many real world applications such as intrusion identification in network systems, detection of ecosystem disturbances and detection of epidemic outbreaks. In this paper, we focus on change point detection in dynamic graphs and address three main challenges associated with this problem: \romannum{1}) how to compare graph snapshots across time, \romannum{2}) how to capture temporal dependencies,  and \romannum{3}) how to combine different views of a temporal graph. To solve the above challenges, we first propose Laplacian Anomaly Detection~(\method) which uses the spectrum of graph Laplacian as the low dimensional embedding of the graph structure  at each snapshot. LAD explicitly models short term and long term dependencies by applying two sliding windows. 
\changed{
Next, we propose \newmethod, a simple and effective generalization of \method to multi-view graphs. \newmethod provides the first change point detection method for multi-view dynamic graphs. It aggregates the singular values of the normalized graph Laplacian from different views through the scalar power mean operation. 
Through extensive synthetic experiments, we show that  \romannum{1}) \method and \newmethod are accurate and outperforms state-of-the-art baselines and their multi-view extensions by a large margin, \romannum{2}) \newmethod’s advantage over contenders significantly increases when additional views are available, and \romannum{3}) \newmethod is highly robust to noise from individual views. In five real world dynamic graphs, we demonstrate that \method and \newmethod identify significant events as top anomalies such as the implementation of government COVID-19 interventions which impacted the population mobility in multi-view traffic networks.}


\end{abstract}




\keywords{Spectral methods, Graph algorithms, Graphs and networks, Machine learning}


\maketitle

\section{Introduction}\label{Sec:intro}

Real world problems in various domains~(e.g. political science, biology, chemistry and sociology) can be modeled as evolving networks that capture temporal relations between nodes. With the increasing
availability of dynamic network data in areas such as social media,
public health and transportation, providing sophisticated methods
that can identify anomalies over time is an important research direction. The task of anomaly detection in dynamic graphs aims to identify different types of time-varying entities that significantly deviate from the ``normal'' or ``expected'' behavior of the underlying graph distribution.

In this work, we focus on change point detection problem which identifies time steps where the graph structure or many graph components deviate significantly from the normal behavior. As change point detection and event detection are closely related, we first explain the distinctions between two of them. Following~\cite{wang2017fast}'s definition, a \emph{change point} is a time point where there is a sudden change in the underlying network generative process and this new process continues beyond the current point. In contrast, an \emph{event} is defined as a time point where the network deviates significantly from the expected behavior and resumes back to normal after this point. Once the anomalous graph instance is found, potential causes can be identified through various static graph analysis techniques. In this work, we first propose a novel change point detection method, Laplacian Anomaly Detection~(\method)~\footnote{our previous work ``Laplacian Change Point Detection for Dynamic Graphs''~\cite{huang2020laplacian} published at KDD 2020}. There are two major challenges for change point detection in dynamic graphs. First, \emph{ how to compare graph snapshots across time.} \method uses the singular values of the Laplacian matrix~(Laplacian spectrum) as a summary for each snapshot because they are closely related to the graph connectivity and low rank approximation. Second, \emph{ how to capture temporal dependencies.} Effective detection of both events and change points require comparison with both recent and distant snapshots. Therefore, \method constructs two context windows which explicitly compare the current graph structure with the typical behaviors from both short term and long term perspectives. 


\begin{figure}[t]
   
    \includegraphics[width=0.6\columnwidth]{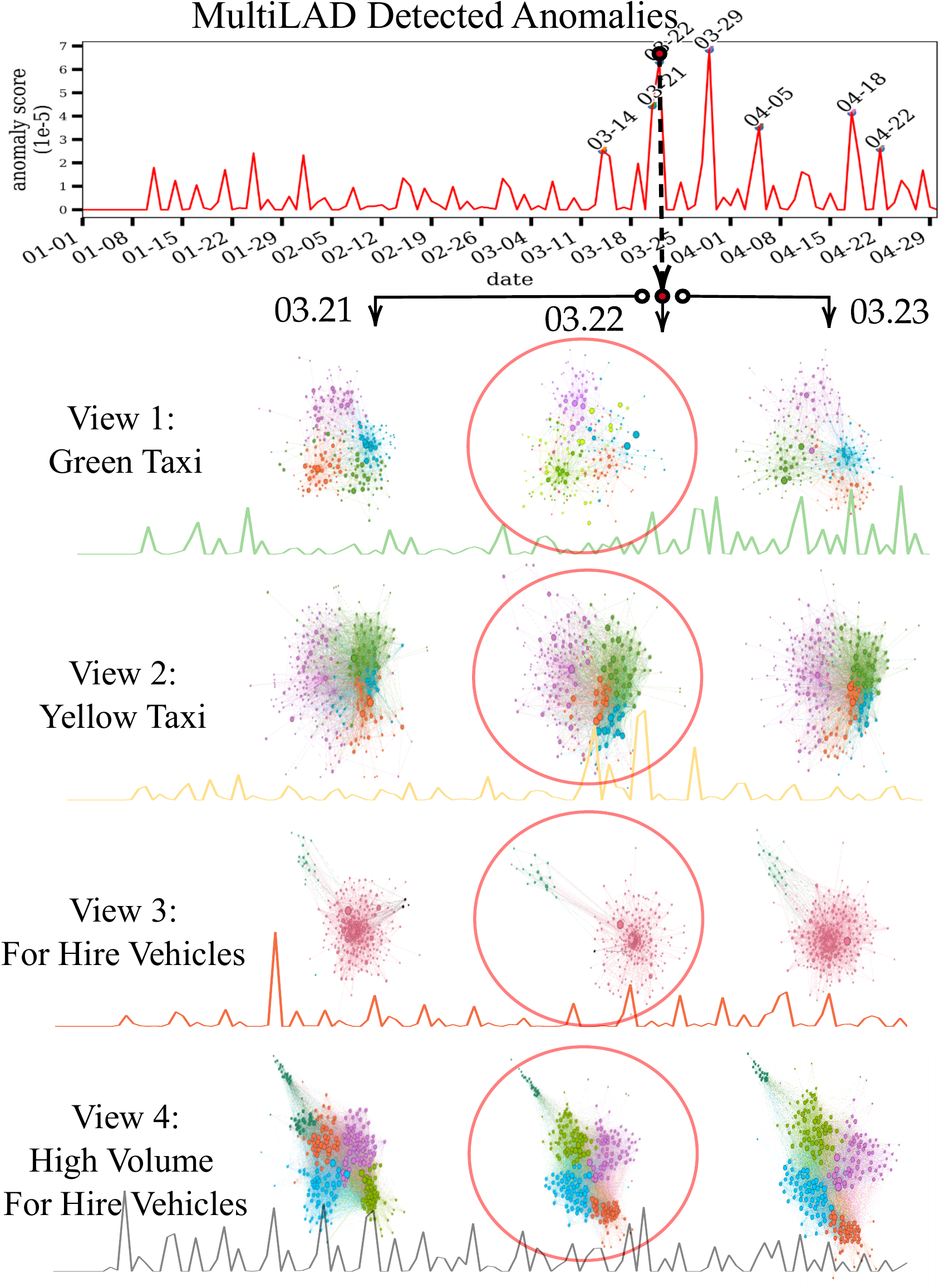}
    \caption{ Our proposed \newmethod accurately detects March 22nd 2020 as an anomalous day in the New York City Transit network. This day is the first day of NYS on Pause Program where all non-essential workers must stay home~\cite{NYCtimeline}. Per each view, we visualize the graph structure on the detected change point as well as a day before and after this point. The anomaly scores from the single-view method \method are also reported for each view to demonstrate the challenge of combining these information when derived independently. On top, we show that \newmethod's anomaly score~(which considers these four views jointly) can effectively reason across views.}
     \label{fig:crown}
     \vskip -0.2in
\end{figure} 

\changed{
In practice, a number of networks can describe the same underlying relations. Multi-view (a.k.a. multiplex) networks are a subset of multi-layer networks which describe relations between the same set of entities from different information sources. Each source can be modelled as an individual view. The interactions between the entities in each view often evolve over time thus forming multi-view dynamic networks. These networks arise naturally in numerous domains such as traffic / transportation networks~\cite{gallotti2015multilayer,SPOTLIGHT}, mobile phone networks~\cite{kiukkonen2010towards}, friendship networks~\cite{eagle2009inferring}, social media networks~\cite{wang2019scg} and citation networks~\cite{tang2009clustering}. Recent work has shown promising improvements by leveraging the additional information from multi-view data in tasks such as community detection~\cite{papalexakis2013more, mercado2018power}, anomalous node detection~\cite{sheng2019multi} and event forecasting~\cite{huang2019mist}. To the best of our knowledge, existing methods studying change point detection only focused on single view dynamic networks~\cite{huang2020laplacian, wang2017fast, ide2004eigenspace, akoglu2010event, peel2015detecting}. Therefore, we propose \newmethod as the first change point detection method for multi-view dynamic networks.
}

\changed{
To better show the motivation for \newmethod, let us consider a real-world application. Given trip records of different taxi companies and for-hire vehicles, can we identify important occasions such as holidays, events or unusual weather conditions which disrupt the overall traffic network? In general, there are two main motivations for leveraging multi-view information for change point detection in dynamic networks. First, data from additional sources may have different characteristics and values~\cite{papalexakis2013more} thus contributing unique and complementary information. In this case, records from taxi companies and for-hire vehicle services~(such as Uber and Lyft) should model the same traffic network but from potentially different age groups and demographics. Second, individual sources can have a high degree of noise and contain disruptive information due to the data collection process. Figure~\ref{fig:crown} shows the anomalous event on March 22nd 2020 in the New York City Transit network detected by our proposed method. This day signals the beginning of the work from home requirement for all non-essential workers. To better illustrate the graph structure, the nodes are colored by their detected community assignment and their sizes correspond to PageRank~\cite{page1999pagerank}. Here, we have four views which are explained in detail in Section~\ref{Sec:real}. Given these four views with different graph structures and evolution, our method effectively reasons across views to find the important anomalies in the underlying traffic network. In contrast, it is often difficult or even infeasible to select one view and apply a single-view change point detection method such as \method~(\method's anomaly scores on each individual views are reported in Figure~\ref{fig:crown}). As seen in the figure, each view can exhibit a different set of anomalies, leading to significant differences among anomalies scores. This disagreement problem is further exacerbated when more views are present. Therefore, naive aggregation strategies on anomaly scores~(such as max or mean) are insufficient, while \newmethod aggregates view-specific features in a meaningful way, preserving the essential information for anomaly detection. After view aggregation, \newmethod utilizes the same anomaly detection procedures as \method thus \newmethod can be seen as a multi-view generalization of \method. 
}

\changed{
Our proposed Multi-view Laplacian Anomaly Detection~(\newmethod) uses the scalar power mean operation to meaningfully aggregate information from individual views and infers the anomaly score based on an aggregated vector. More specifically, \newmethod first computes the singular values of the normalized Laplacian matrix from each view as the view specific feature vector. Then, \newmethod merges information across different views by aggregating these feature vectors through the scalar power mean operation. In this way, the effect of noise from individual views are lessened while key information about the overall generative process of the multi-view network is preserved. Lastly, \newmethod compares the aggregated vector with past normal behavior and determines if a particular time step is anomalous or not.
}

\medskip
\noindent\textbf{Summary of contributions}: 
\begin{itemize}[topsep=0pt]
    \item We introduce a novel change point detection method: Laplacian Anomaly Detection~(LAD). LAD utilizes the singular values of the Laplacian matrix to obtain a low dimension embedding of graph snapshots. To the best of our knowledge, this is the first time that the Laplacian spectrum has been used for change point detection.
    \item \method explicitly captures both the short term and long term temporal relations to detect both events and change points.  
    \item We extensively evaluate \method on three synthetic tasks and three real world datasets. We show that \method is more effective at identifying significant events than state-of-the-art methods. 
    \item \changed{ We extend \method to the multi-view setting by proposing \newmethod: the first change point detection method for multi-view dynamic networks. \newmethod aggregates the singular values of the normalized Laplacian matrices through scalar power mean and identifies the most informative singular values from each view.}
    \item \changed{We extensively evaluate \newmethod on synthetic experiments and show that \newmethod 1). gains increased performance from additional views, 2). is highly robust to noise, and  3). significantly outperforms state-of-the-art single view baselines and their  multi-view extensions.}
    \item \changed{We apply \newmethod to two real world multi-view traffic networks. We demonstrate that \newmethod correctly detects major real world traffic disruptions such as the implementation of stay-at-home order for COVID-19 and Christmas day.}
\end{itemize}

\medskip
\noindent\textbf{Reproducibility}: All code is publicly available. The code for \method and \newmethod are available at \\
\url{https://github.com/shenyangHuang/LAD.git} and \url{https://github.com/shenyangHuang/multiLAD.git} .

\section{Related Work}



\changed{
The common strategy across most anomaly detection methods is to extract a low dimensional representation from graph snapshots and then apply an anomaly scoring function to compare these representations~\cite{ranshous2015anomaly}. In this section, we discuss related literature from both single-view and multi-view anomaly detection. We compare features of \method, \newmethod and other alternative methods in Table~\ref{tab:salesman}. Note that \newmethod is the only approach that satisfies all the desired properties and is designed for multi-view dynamic graphs. 
}

\subsection{Event Detection}
An early work by Idé and Kashima~\cite{ide2004eigenspace} examines runtime anomalies at the application layer of multi-node computer systems. A web-based system is modeled as a weighted graph where each edge represents a dependency between services. Idé and Kashima aim to find time points where the majority of the edge attributes in the network show significant deviation from the recent ones. The principal eigenvector corresponding to the maximum eigenvalue of the positive weighted adjacency matrix $W$ is used as a low dimensional representation of the graph~(called \emph{activity vector}). Activity vectors from a short term context window form a matrix upon which singular value decomposition~(SVD) is then performed. The typical graph behavior within the context window is then represented as the principle left singular vector. The deviation of the current activity vector from the typical behavior is used as the anomaly score. Different from Idé and Kashima, we use the Laplacian spectrum to summarize graph structures and explicitly constructs two sliding windows for long term and short term contexts. 

In~\cite{koutra2012tensorsplat}, a dynamic graph is viewed as a high order tensor. The authors proposed to perform the PARAFAC decomposition~\cite{bro1997parafac} of this tensor to obtain the signature vector for anomaly detection. As a multi-view dynamic graph can be considered as a 4th order tensor, we generalize TENSORSPLAT to the multi-view setting, and include it as one of the baselines in our experiments in Section~\ref{sec:multiview}.

SPOTLIGHT~\cite{SPOTLIGHT} was proposed to detect anomalies in dynamic graphs and is centered around monitoring the (dis)appearance of large or dense subgraphs in a dynamic graph. SPOTLIGHT focuses on the related but different anomalous \emph{subgraph} detection task, while we detect anomalous graph snapshots. In this work, we also investigate the New York City Transit 2015-2016 dataset studied in~\cite{SPOTLIGHT}.

\begin{table}[tp]  
\begin{center}
\begin{tabular}{ l|c|c|c|c||c|}
       \diagbox{Property}{Method} &  \rotatebox{90}{Activity vector~\cite{ide2004eigenspace}} & \rotatebox{90}{TENSORSPLAT~\cite{koutra2012tensorsplat}} & \rotatebox{90}{EdgeMonitoring~\cite{wang2017fast} } & \rotatebox{90}{LAD~\cite{huang2020laplacian}} & \rotatebox{90}{\newmethod} \\ 
    \hline 
    multi-view          &                & \CheckmarkBold   &                &                  & \CheckmarkBold \\
    robust to noise     &                &                  &                &                  & \CheckmarkBold\\ 
	event               & \CheckmarkBold & \CheckmarkBold  &                & \CheckmarkBold   & \CheckmarkBold \\ 
    change point        &                &                  & \CheckmarkBold & \CheckmarkBold   & \CheckmarkBold \\
    evolving \# nodes   & \CheckmarkBold &                  &                & \CheckmarkBold   & \CheckmarkBold \\ 
    node permutation invariant & \CheckmarkBold & \CheckmarkBold & & \CheckmarkBold & \CheckmarkBold  \\ 
   \hline
\end{tabular}
\caption{Only \newmethod meets all desirable properties.\label{tab:salesman}}
\end{center}
\vskip -0.3in
\end{table}



\subsection{Change Point Detection}


Recently, \citet{wang2017fast} model network evolution as a first order Markov process and propose the EdgeMonitoring method based on Monte-Carlo sampling techniques. Their assumption is that there is some unknown underlying model that governs the generative process. Moreover, each graph snapshot is dependent on the current generative model as well as the previously observed snapshot. This method is often regarded as the current state-of-the-art for change point detection. However, EdgeMonitoring relies on consistent node orderings across all time steps and assumes constant number of nodes for each snapshot. This assumption is easily violated in large social networks where users accounts are added frequently. In contrast, \method can manage varying number of nodes across time.


\subsection{Multi-view Anomaly Detection}
\changed{
Most multi-view anomaly detection methods follow a two step procedure: 1). extract view-specific representations and 2). aggregate such representations to identify anomalous data. While prior work~\cite{teng2017anomaly,sheng2019multi,guo2018partial,sohn2017bayesian} focused on the anomalous node~(or instance) detection task, we tackle the change point detection problem in the multi-view \textit{dynamic} graph setting. In other words, instead of detecting anomalous nodes, we spot anomalous timestamps when the structure of the entire graph changes.
}

\changed{
We draw on the same idea used in node-level anomaly detection methods which assumes that abnormal events would create a disturbance of regularities across all views. By mining such irregular patterns, one can achieve stronger anomaly detection results than with just a single view. In this work, we use this idea to identify anomalous graph snapshots across different views to detect significant changes in the overall graph generative process for all views.
}

\subsection{Multi-view Clustering}
\changed{
The goal of graph clustering is to obtain groups of nodes that are similar with regards to some structural or node attribute information~\cite{papalexakis2013more,tang2010community}. Graph clustering in the multi-view setting is a similar task to multi-view change point detection in which one needs to combine information from different sources on the same nodes to extract some underlying structure. However, multi-view graph clustering focuses on static graphs while in this work we consider dynamic graphs.
}

\changed{
Mercado \etc~\cite{mercado2018power} propose to take the matrix power means of Laplacians~(PML) to extend the spectral clustering algorithm to multilayer graphs. Inspired by the effectiveness of power mean operations in multi-view clustering, we propose to utilize the scalar power mean operation to merge information from different views for change point detection. 
}

\changed{
Sohn \etc~\cite{sohn2017bayesian} proposed a Bayesian multilayer stochastic blockmodeling framework for multilayer clustering. They also extend their method to detect communities changes in multilayer temporal graphs. However, their approach is limited to networks with block structures. In comparison, \newmethod is applicable to any graph including those with no known community structures~(such as the Barabási-Albert model).}

\section{Problem definition}
In this section, we formally define the multi-view change point and event detection problems which encompasses the classical single-view definition of the same problems as a special case.

\textbf{Multi-view Dynamic Graph}
\changed{
A multi-view dynamic graph  $\mathbb{G} = \{ \mathcal{G}_t \}_{t=1}^{T}$ is given by a collection of graph snapshots $\mathcal{G}_t = \{ \mathbf{G}_{t,r} \}_{r=1}^m$, where $\mathbf{G}_{t,r}$ is the graph from the view (or relation) $r$ at time step $t$. At each time step, each graph $\mathbf{G}_{t,r} = (\mathbf{V}, \mathbf{E}_{t,r})$ consists of the set of nodes $\mathbf{V}$ and a set of edges $\mathbf{E}_{t,r} \subset \mathbf{V} \times \mathbf{V}$.  For clarity of exposition, we assume that the set of nodes is fixed over time and each view has the same number of nodes~(though \newmethod can handle settings where this assumption does not hold, see Section~\ref{sub:adapt}). Each edge $e=(i,j,w) \in \mathbf{E}_{t,r}$ is then defined as the connection between node $i$ and node $j$ at timestamp $t$ in the view $l$ with weight $w \in \mathbb{R}^{+}$. Note that in practice both  nodes and edges can appear or disappear from one time step to another.
We denote by $\mathbf{A}_{t,r} \in \mathbb{R}^{n \times n}$ the adjacency matrix representing edges in $\mathbf{E}_{t,r}$ where $n = |\mathbf{V}|$. When the number of views $m=1$, we fall back into  the single-view setting similar to~\cite{wang2017fast,ide2004eigenspace}.
}

\textbf{Multi-view Change Point Detection}
\changed{
Based on the above definition, let $\mathbb{G}$ be a multi-view dynamic graph and let $\mat{H}_t$ be the underlying graph generative model at time $t$. $\mat{H}_t$ is not observed but is assumed to control the graph behavior across all views. }
\changed{
The goal of multi-view change point detection is to
find time points after which $\mat{H}_t$ significantly differs from the previous steps. More precisely, we want to find a set $S \subseteq \{1, ..., T\}$ such that for each $t \in S$,  $ \cdots \simeq \mat{H}_{t-2} \simeq \mat{H}_{t-1} \not \simeq \mat{H}_{t} \simeq \mat{H}_{t+1} \simeq \cdots$. 
}


\textbf{Multi-view Event Detection}
\changed{
In this work, we also consider events similar to~\cite{ide2004eigenspace}. Let $\mathbb{G}$ be a multi-view dynamic graph and let $\mat{H}_t$ be the underlying graph generative model at time $t$. The goal of event detection is to find a set of time points $S \subseteq \{1, ..., T\}$ such that for each $t \in S$, $ \mat{H}_{t-1} \not \simeq \mat{H}_{t}$ and $ \mat{H}_{t-1} \simeq \mat{H}_{t+1}$. Events are one time sudden changes in graph structure where the generative model $\mat{H}_{t+1}$ reverts back to normal after the event at time $t$. 
}



\section{Laplacian Anomaly Detection}\label{sec:LAD}
We first propose a new spectral anomaly detection method for single-view dynamic graphs: Laplacian Anomaly Detection~(LAD).
The core idea of \method is to detect high level graph changes from low dimensional embeddings~(called signature vectors). The ``typical'' or ``normal'' behavior of the graph can then be extracted from a stream of signature vectors based on both short term and long term dependencies. In this way, we can compare the deviation of the current signature vector from the normal behavior. 

\subsection{Laplacian Spectrum}

\begin{figure}[t]
    \begin{center}
        \centerline{\includegraphics[width=0.7\columnwidth]{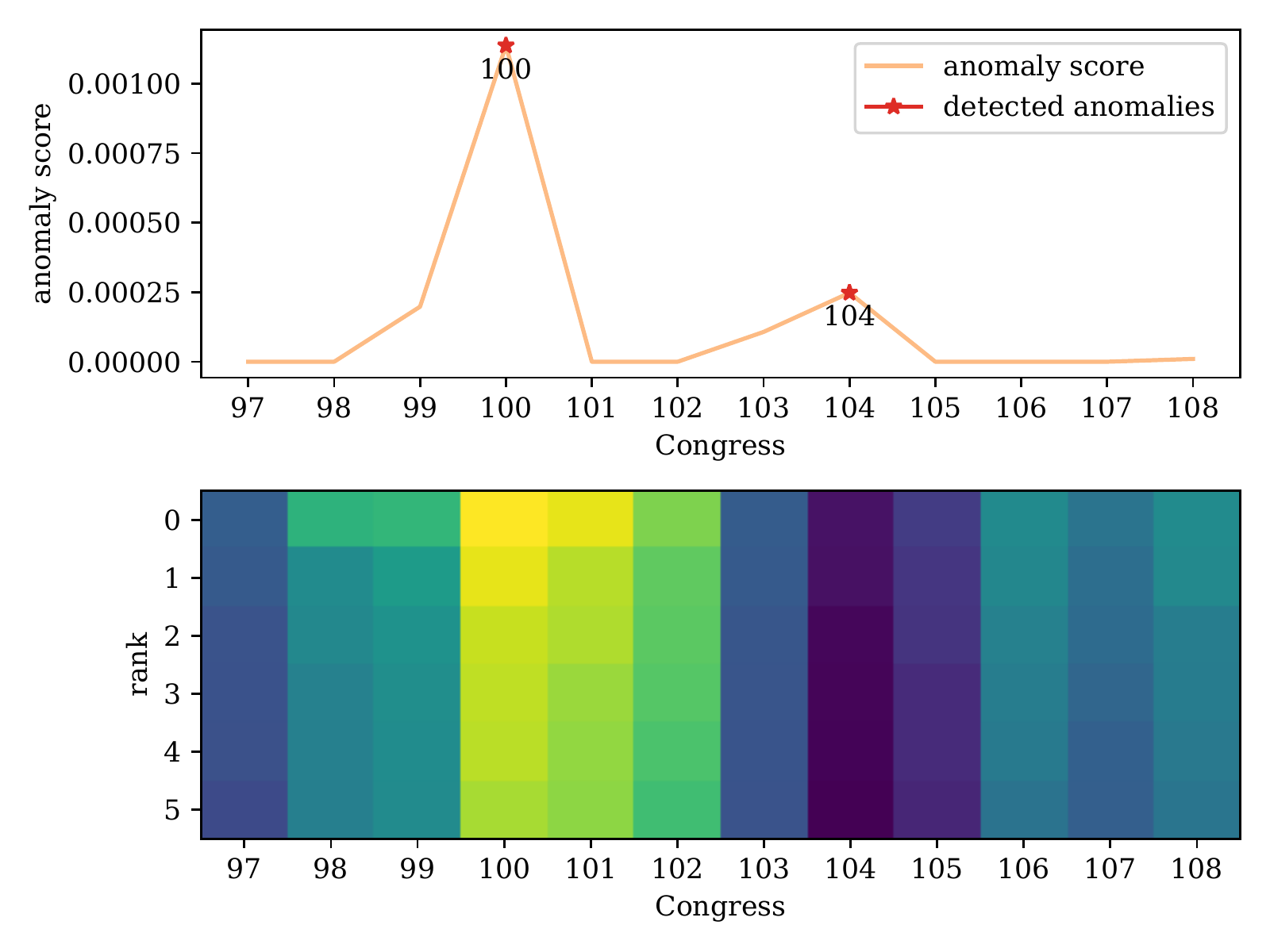}}
        \vspace{-3pt}
        \caption{\method captures changes in the graph spectrum. \method scores (top) and the top 6 singular values (bottom) are aligned and show correspondence at each time step. This illustration is from the Senate co-sponsorship network. The warmer the color, the higher the intensity.}        
        \label{fig:specSenate}
    \end{center}
    \vskip -0.3in
\end{figure}

We first define the~(unnormalized) Laplacian matrix $\mat{L}$ as $\mat{L} = \mat{D} - \mat{A}$ where $\mat{D}$ is the diagonal degree matrix and $\mat{A}$ is the adjacency matrix of $\mathbf{G}$. Here we drop the subscripts for readability, $\mat{L}$ is defined based on a given view at a given timestep. For \method, we use the vector of singular values obtained through Singular Value Decomposition~(SVD)~\cite{golub1971singular} of $\mat{L}$ as graph embedding for a snapshot $\mathbf{G}$. Figure~\ref{fig:specSenate} shows the visualization of the Laplacian spectrum and the corresponding anomaly scores detected by \method for the Senate co-sponsorship network. This choice is motivated by the fact that singular values
\begin{itemize}
    \item are related to the Laplacian spectrum which is fundamentally related to structural properties of the graph such as community structure and degree distribution,
    \item encodes the compression loss of a low rank approximation of the Laplacian matrix,
    \item are node permutation invariant,
    \item can be efficiently computed for real world sparse matrices.
\end{itemize}

First, it is known that the singular values of a positive symmetric matrix coincides with its eigenvalues. The Laplacian matrix is symmetric and positive semi-definite for an undirected weighted graph~\cite{von2007tutorial}. The eigenvalues of the Laplacian matrix capture fundamental structural properties of the corresponding graph. These properties have been extensively utilized in many fields such as randomized algorithms, combinatorial optimization problems and machine learning~\cite{zhang2011laplacian}. Notably, the field of spectral graph theory~\cite{chung1997spectral} extensively studies properties of the graph Laplacian matrix. As an illustration, the multiplicity $k$ of the eigenvalue $0$ of $\mat{L}$ is equal to the number of connected components of the graph~\cite{von2007tutorial}. In addition, the Laplacian spectrum is related to many other structural properties of the graph such as the degree sequence, number of connected components, diameter, vertex connectivity and more~\cite{zhang2011laplacian}.

Second, it is well known that the truncated SVD gives the best low rank approximation of a matrix with respect to both the Frobenius norm and the 2-norm (Eckart-Young theorem). More precisely, the $(k+1)$th singular value $\sigma_{k+1}$ of a matrix corresponds to the reconstruction error of the best rank $k$ approximation measured in the 2-norm. From this insight, we know that the~(ordered) singular spectrum $\sigma_1, \sigma_2, ..., \sigma_r$ encodes rich information regarding the reconstruction loss that would occur for different levels of low rank approximations. Truncated SVD is often used as a powerful compression tool for images~\cite{rufai2014lossy}, videos~\cite{benjamin2017compressed} and audio~\cite{zamani2017frequency}. Thus, capturing the singular spectrum can be seen as an alternative compression based anomaly detection technique as categorized by Ranshous et al.~\cite{ranshous2015anomaly}. Intuitively, huge fluctuations in the singular values of the Laplacian matrix reflect drastic changes to the global graph structure. 

Third, being node permutation invariant is one of the desirable properties for a graph learning method~\cite{xu2018powerful}. \method is permutation invariant as singular values of the Laplacian matrix is invariant to row and column permutations. Lastly, singular values can be efficiently obtained through sparse computations~\cite{aharon2006k} and we discuss the computational complexity of \method in Section~\ref{sec:CC}.


\subsection{Characterizing Normal Behavior} \label{LAD:normal}
Identifying a normal or typical behavior from a sliding window is often an integral part of change point detection. Similar to~\cite{akoglu2010event, ide2004eigenspace}, we compute a ``typical'' or ``normal'' behavior vector from the Laplacian spectrum of the previous $l$ time steps, where $l$ is the window size. First, we perform \emph{L2} normalization on the Laplacian spectrum seen so far $\vec{\sigma}_0, \hdots, \vec{\sigma}_t\in\mathbb{R}^n$ to obtain unit vectors. Next, a context matrix $\mat{C}$ is constructed:
\begin{equation}\label{eq:def.C}
    \mat{C} = \left( \begin{array}{cccc}
                 |& | & &|  \\
                 \vec{\sigma}_{t-l} & \vec{\sigma}_{t-l+1} & \hdots & \vec{\sigma}_{t-1} \\ 
                 |& | & & |
            \end{array} \right) \in \mathbb{R}^{n \times l}
\end{equation}
where $n$ is the length of the signature vector~(which corresponds to the number of nodes in the graph). We compute the left singular vector of $\mat{C}$ with SVD to obtain the normal behavior vector $\tilde{\sigma}_t$. One can interpret this behavior vector as a weighted average of the sliding window spectrums~\cite{akoglu2010event}. 

We propose to compare the current signature vector with the typical behavior from two independent sliding windows: a short term window and a long term window. The short term window encodes information from the most recent trend and captures abrupt changes in the overall graph structure. Depending on the application, the length of the short term window can be adjusted to best reflect an appropriate time scale. On the other hand, a long term window is designed to capture larger scale and more gradual trends in the dynamic graph. For example, for the UCI Message dataset, the short term context can be monitoring weekly change and the long term context can be monitoring a biweekly change.

\subsection{Anomalous Score Computation}
After capturing the normal behavior, a scoring function is then defined to measure the difference between the current signature vector and the expected or normal one. In this work, we use the same anomaly score as introduced in~\cite{ide2004eigenspace, akoglu2010event}, namely the $Z$ score. Let $\tilde{\sigma}_t$ be the normal behavior vector and $\vec{\sigma}_t$ be the Laplacian spectrum at current step. As mentioned in Section~\ref{LAD:normal}, both $\tilde{\sigma}_t$ and $\vec{\sigma}_t$ are normalized to unit vectors, then the $Z$ score is computed as:
\begin{equation}
\label{Eq:Zscore}
    Z = 1 - \frac{\vec{\sigma}_t^\top \tilde{\sigma}_t }{ \|\vec{\sigma}_t\|_2 \|\tilde{\sigma}_t\|_2} = 1 - \vec{\sigma}_t^\top \tilde{\sigma}_t = 1 - \cos{\theta},
\end{equation}
where $\cos{\theta}$ is the cosine similarity between $\vec{u}$ and $\vec{v}$. Essentially the $Z$ scores becomes closer to 1 when the current spectrum is very dissimilar to the normal thus signaling an anomalous point. 

Let $w_s$ and $w_l$ be the sizes of the short term and long term sliding windows respectively. Analogous to Section~\ref{LAD:normal}, one can obtain two different normal behavior vectors based on the two sliding windows. One can then compute the short term and long term anomaly scores $Z_{s}$ and $Z_l$ based on cosine similarity. To best aggregate these two perspectives, we take  $Z_t = max(Z_s, Z_l)$ to decide if the current graph is more anomalous in abrupt or gradual changes. 

$Z_t$ scores can be used to detect events (one time changes in the graph) while change points are often followed by a sequence of decaying anomaly scores. We aim to detect both events and change points thus we compute the \emph{final} anomaly score $Z_t^* = min(Z_t - Z_{t-1},0)$ which emphasizes the time points with the largest increase in $Z_t$ when compared to $Z_{t-1}$. Additional discussion and visualization are added in Appendix~\ref{app:zscore}.


\section{\newmethod}
\begin{algorithm}[t]
    \caption{\newmethod}
    \label{algo:flow}
    \SetAlgoLined
    \KwIn{Multi-view graph $\mathbb{G}$}
    \Parameter{Power $p$, sliding window sizes $w_s,w_l$, embedding size $k$}
    \KwOut{Final anomaly scores $Z^*$}
    \BlankLine
    \ForEach{multi-view snapshot $\mathcal{G}_t$ in the multi-view graph $\mathbb{G}$}{
        \ForEach{graph snapshot $\mathbf{G}_{t,r} \in \mathcal{G}_t$}{
        Compute $\mat{L}_{sym}$ (see Eq.~\eqref{eq:def.sym.laplacian})\;
        Compute top $k$ singular values $\tilde{\sigma}_{t,r}$ of  $\mat{L}_{sym}$\;
        }
        Let $\vec{\Sigma}_{t} = s_p(\tilde{\sigma}_{t,1}, \hdots, \tilde{\sigma}_{t,r})$ \;
        Perform L2 normalization on $\vec{\Sigma}_{t}$\;
        Compute left singular vector $\tilde{\Sigma_{t}^{w_s}}$ of context $\mat{C}_{t}^{w_s} \in \mathbb{R}^{k \times w_s}$ (see Eq.~\eqref{eq:def.C})\;
        Compute left singular vector $\tilde{\Sigma_{t}^{w_l}}$ of context $\mat{C}_{t}^{w_l} \in \mathbb{R}^{k \times w_l}$(see Eq.~\eqref{eq:def.C})\;
        $Z_{t}^{w_s} = 1 - \vec{\Sigma}_t^\top \tilde{\Sigma}_{t}^{w_s}$ \;
        $Z_{t}^{w_l} = 1 - \vec{\Sigma}_t^\top \tilde{\Sigma}_{t}^{w_l}$ \;
    }
    \ForEach{time step $t$}{
    $Z_{s,t}^* = max(Z_{w_s,t} - Z_{w_s,t-1},0)$\;
    $Z_{l,t}^* = max(Z_{w_l,t} - Z_{w_l,t-1},0)$\;
    $Z_{t}^* = max(Z_{s,t}^*,Z_{l,t}^*)$\;
    }
    Return $Z^*$\;
\end{algorithm}

\changed{
In this section, we present \newmethod  for multi-view change point detection in dynamic graphs. The overall procedure for \newmethod is summarized in Algorithm~\ref{algo:flow}.
}

\subsection{Extracting Signature Vectors Per View}

\changed{
Using the unnormalized Laplacian matrix $\mat{L}$ can be problematic in the multi-view setting where edge weights from different views may vary significantly in magnitude. Therefore, we use the symmetric normalized Laplacian~\cite{von2007tutorial} instead:}

\begin{equation}\label{eq:def.sym.laplacian}
    \mat{L}_{sym} = \mat{D}^{-\frac{1}{2}} \mat{L} \mat{D}^{-\frac{1}{2}} = \mat{I} - \mat{D}^{-\frac{1}{2}} \mat{A} \mat{D}^{-\frac{1}{2}}
\end{equation}
%
\changed{
In this way, the Laplacian matrix of each view is normalized by node degree while also being symmetric (the eigenvalues are the same as singular values). Note that the singular values $\lambda_i$ of $\mat{L}_{sym}$ are now bounded in $[0,2]$~(in contrast with the singular values $\sigma_i$ of the un-nromalized Laplacian which were  unbounded). We consider the singular values vector $\pmb{\vec{\lambda}}_{r,t} = [\lambda_{1}, \dots, \lambda_{n}]$ of the normalized Laplacian $\mat{L}_{sym}$ as the signature vector for the graph $\mathbf{G}_{t,r}$  of the $r$-th view at time step $t$ . Therefore, at each time step $t$, we obtain $m$ signature vectors coming from each view. 
}

\subsection{Aggregating Per View Signatures}
\changed{
To aggregate the per-view signature vectors for multi-view change point detection, we propose to use the \emph{scalar power mean} operation. The $p$th order power mean of a set of non-negative real numbers $x_1, \hdots, x_m$ is defined by:}
\begin{equation}\label{eq:def.spm}
    s_p(x_1, \hdots, x_m) = \left(\frac{1}{m} \sum_{i=1}^{m} x_i^p\right)^{\frac{1}{p}}
\end{equation}
\changed{
where $p \in \mathbb{R}$ is an hyper-parameter. 
We denote the vector of singular values of the Normalized Laplacian matrix $\mat{L}_{sym}$ in view $r$ at time $t$ by $\pmb{\vec{\lambda}}_{r,t}$ as before. 
We now define the \emph{scalar power mean spectrum} $\Lambda_t$ of a multi-view dynamic graph with $m$ views at time $t$ as:}
\begin{equation}\label{eq:def.Sigma_S}
    \Lambda_t = s_p(\pmb{\vec{\lambda}}_{1,t}, \dots, \pmb{\vec{\lambda}}_{m,t}) \in \mathbb{R}^n
\end{equation}
\changed{
where $s_p$ is applied component-wise to the singular value vectors, i.e., $(\Lambda_{t})_i = s_p((\pmb{ \vec{\lambda}}_{1,t})_i, \dots, (\pmb{\lambda}_{m,t})_i)$ for all $0 \leq i \leq n$. In this way, $\Lambda_t \in \mathbb{R}^n$ is aggregated from the set of per-view signature vectors.
}

\changed{
In this work, we set $p=-10$ as empirically it shows the best performance. This is also motivated by the fact that the smallest eigenvalues of the Laplacian contain crucial information on the structure of a graph and a negative power magnifies the weight of smaller eigenvalues during aggregation. An Ablation study on the choice of $p$ is discussed in Appendix~\ref{app:power}. We also add a small diagonal shift to the normalized Laplacian $\mat{L}_{sym}$ to ensure that it is positive definite. For all experiments, we set the shift $\epsilon = log(1+|p|)$ for $p<0$ thus no $0$ would be encountered in Equation~\eqref{eq:def.spm}.}


\section{Method Properties}

\subsection{Evolving Graph Size and Node Permutation} \label{sub:adapt}

\changed{
For clarity, we assumed that the number of nodes in multi-view dynamic graphs is constant over time and the same across all views. In many real world graphs, the number of nodes often change over time. In practice, \method is able to adapt to the evolving size of dynamic graphs by computing the top $k$ singular values as signature vectors instead of the full spectrum. $k$ can be chosen to be the number of nodes of the smallest graph snapshot seen so far. Similarly, \newmethod can be applied to multi-view dynamic graphs where the graph size changes by considering the top $k$ singular values from each view before aggregation. In addition, \newmethod can also adapt to multi-view graphs where the number of nodes in each view differs from each other, again by considering the top $k$ singular values where $k$ is the size of the view with the least number of nodes. }

\changed{
Since the Laplacian spectrum is invariant to row/column permutations, both \method and \newmethod are invariant to node permutations. This implies that the nodes of a dynamic graph are not required to conserve the same ordering for each graph snapshot. This can be very relevant for privacy sensitive applications where the alignment of nodes across many time steps could cause information leakage. This property can also be very useful in the multi-view setting where aligning  node order across views may be technically challenging. In this way, \method and \newmethod can be applied to a broader range of scenarios than methods relying on a consistent node ordering across snapshots. }

\subsection{Computational Complexity}\label{sec:CC}
\changed{
The computational complexity of \method is dominated by the singular value decomposition of the Laplacian, which in general has complexity $\mathcal{O}(n^3)$ for the Laplacian matrix $\mat{L} \in \mathbb{R}^{n \times n}$. As many real world networks are sparse, sparse SVD can be used to reduce the complexity to $\mathcal{O}(k|\mat{E}|)$, where $k$ is the number of singular values and $|\mat{E}|$ is the number of edges in the graph. In a multi-view dynamic graph with $T$ time steps, $m$ views and an average of $|\mat{E}|$ edges per snapshot, \newmethod has a complexity of $\mathcal{O}(k \cdot m \cdot T \cdot |\mat{E}|)$ when computing $k$ singular values. Note that the scalar power mean aggregation is very fast to compute thus computation cost of \newmethod lies mainly in computing SVD for each view. }


\definecolor{lightblue}{rgb}{0.9,0.98,1}
\definecolor{lightgreen}{rgb}{.9,1,0.95}

\section{Single-View Experiments} \label{SV:dataset}
In this section, we conduct extensive experiments on both synthetic and real world single-view dynamic graphs to evaluate the performance of \method.

\subsection{Measuring Performance}
To quantitatively evaluate the quality of detected change points, we choose the Hits at $n$~($H@n$) metric which reports the number of correctly identified anomalies out of the top $n$ time points with the highest anomaly scores.
In synthetic experiments, we use the planted anomalies in the generation process for evaluation. 
In real world datasets, we treat the well-known anomalous time steps from the literature as ground truth anomalies.


\subsection{Contenders and Baselines}
We compare \method with alternative single-view methods. We use the same sliding window sizes across all methods when applicable. Both the anomalies scores from the short term and long term sliding windows are shown in Figure~\ref{fig:SVeventchange},\ref{fig:UCIscore},\ref{fig:senateScore},\ref{fig:canVotescore}.
Note that for all experiments, the startup period~($0,\hdots,l$) is set to have anomaly score of 0 because we assume these points are not change points.  
\begin{itemize}[leftmargin=12pt]
    \item \textbf{Activity vector.} We refer to the method proposed by Idé et al.~\cite{ide2004eigenspace} as ``activity vector'' based methods. Idé et al. used the principal eigenvector of the adjacency matrix~(namely the activity vector) instead of our proposed Laplacian spectrum. According to the original work, only a short term context window is considered. 
    
    \item \textbf{TENSORSPLAT.} Koutra et al.~\cite{koutra2012tensorsplat} proposed to view the temporal graph as a tensor and then perform the PARAFAC or CP decomposition to obtain low dimensional factors which group similar entities or timestamps together~(we use CP rank of 30 for all experiments). The original paper proposed to use clustering on the factors for change detection. However, the clustering algorithm is not specified. We use the well-known Local Outlier Factor~(LOF)~\cite{breunig2000lof} approach along with the TENSORSPLAT framework.
    
    \item \textbf{EdgeMonitoring.} Wang et al.~\cite{wang2017fast} proposed the EdgeMonitoring approach and used joint edge probabilities as the feature vector while modelling network evolution as a first order Markov process. 
\end{itemize}

\subsection{Synthetic Experiments}

\begin{table*}[t]
\caption{Experiment Setting: the changes in the generative model in \\
(a) SBM experiment~(Sec.~\ref{sec:pure}) when only the number of blocks ($N_c$) changes (Pure setting). The more communities there are, the higher the color intensity is. \\
(b) SBM experiment~(Sec.\ref{sec:hybrid}) where we have a combination of event and change points (Hyprid setting).\\} \label{tab:SVSBM}
\begin{subtable}[t]{0.3\columnwidth}
\begin{tabular}{c | l | c | c }
\toprule
\multicolumn{4}{c}{Pure Setting} \\
\midrule
Time& $N_c$ & $p_{in}$ & $p_{ex}$ \\
\midrule

\rowcolor{blue!4} 0 & 4 & 0.25 & 0.05 \\ 
\rowcolor{blue!10} 16 & 10 & 0.25 & 0.05 \\ 
\rowcolor{blue!2} 31 & 2 & 0.5 & 0.05 \\ 
\rowcolor{blue!4} 61 & 4 & 0.25 & 0.05 \\ 
\rowcolor{blue!10} 76 & 10 & 0.25 & 0.05 \\ 
\rowcolor{blue!2} 91 & 2 & 0.5 & 0.05 \\ 
\rowcolor{blue!4} 106 & 4 & 0.25 & 0.05 \\
\rowcolor{blue!10} 136 & 10 & 0.25 & 0.05 \\
\bottomrule
\end{tabular}
\caption{anomalies in Sec.~\ref{sec:pure}: Pure setting}
\end{subtable}%
\begin{subtable}[t]{0.35\columnwidth}

\begin{tabular}{c | l | c | c | c}
\toprule
\multicolumn{5}{c}{Hybrid Setting}\\
\midrule
Time & Type &  $N_c$ & $p_{in}$ & $p_{ex}$\\
\midrule
0 & start point & 4 & 0.25 & 0.05 \\ \rowcolor{lightgreen}
16 & event & 4 & 0.25 & 0.15 \\ \rowcolor{blue!5}
31 & change point & 10 & 0.25 & 0.05 \\ \rowcolor{lightgreen}
61 & event & 10 & 0.25 & 0.15 \\ \rowcolor{blue!5}
76 & change point & 2 & 0.5 & 0.05 \\ \rowcolor{lightgreen}
91 & event & 2 & 0.5 & 0.15\\ \rowcolor{blue!5}
106 & change point & 4 & 0.25 & 0.05 \\ \rowcolor{lightgreen}
136 & event & 4 & 0.25 & 0.15 \\ 
\bottomrule
\end{tabular}
\caption{anomalies in Sec.~\ref{sec:hybrid}: Hyprid setting}
\end{subtable}%
\end{table*}


To demonstrate the performance of \method,  we design three controlled experiments. For these synthetic experiments, we use data generated from the Stochastic Block Model (SBM)~\cite{holland1983stochastic}. The number of communities $N_c$ as well as the number of nodes within each community can be specified through a size vector $\vec{c}\in \mathbb{R}^{k}$. In addition, the inter-community and intra-community connectivity for each block can be directly encoded in a symmetric probability matrix $P \in \mathbb{R}^{N_c \times N_c}$. For all experiments, we set the short term and the long term window to be 5 and 10 time steps respectively and use the entire Laplacian spectrum for \method. 

\subsubsection{\textbf{Pure Setting}} \label{sec:pure}
Here, we only introduce change points, where the adjustments in community structure persists until the next change point is reached. We generate a temporal network with 151 time points where each snapshot is produced through SBM parametrized by $\vec{c}$ and $P$. There are always 500 nodes per snapshot and the community changes are described in Table~\ref{tab:SVSBM}. Here $N_{c}$ represents the number of equal sized communities in the snapshot, and $p_{in}, p_{ex}$ denotes the internal and external community connectivity probability respectively. We set the continuity rate~\cite{wang2017fast} to be 0 for change points and 1.0 elsewhere for the pure setting. 

The top 7 most anomalous points correspond to the 7 ground truth points in Table~\ref{tab:SVSBM} while the other points have extremely low anomaly scores. In this experiment, EdgeMonitoring and \method both achieves perfect precision as both can reason with gradual changes over time. In contrast, both TENSORSPLAT and Activity vector can only recover some change points. As Activity vector uses the principal eigenvector of the adjacency matrix, it is unable to detect community changes as effectively as the Laplacian spectrum used in \method.

\subsubsection{\textbf{Hybrid Setting}}
\label{sec:hybrid}

\begin{figure}[t]
    \begin{center}
        \centerline{\includegraphics[width=0.7\columnwidth]{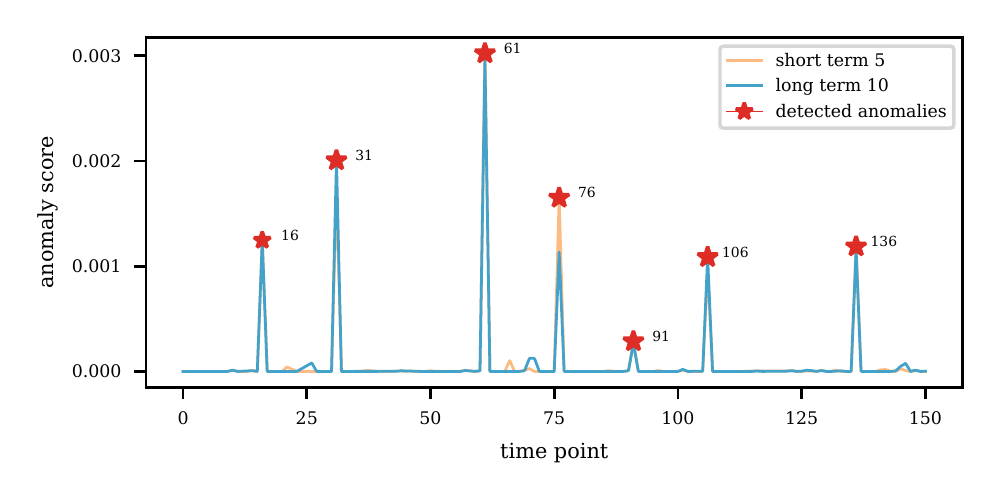}}    
        \caption{\method perfectly recovers all \underline{events} and \underline{change} points defined in Table~\ref{tab:SVSBM}.}
        \label{fig:SVeventchange}
    \end{center}
    \vskip -0.3in
\end{figure}

To study the effectiveness of \method in detecting both change points and events, we generate a synthetic experiment with SBM where these two types of anomalies both exist. We generate events by strengthening the inter-community connectivity for a given time point~(subsequent points are not affected). These events can correspond to a sudden increase in collaborations between typically separated communities such as political parties. The details regarding the generative process can be seen in Table~\ref{tab:SVSBM}. We set the continuity rate~\cite{wang2017fast} to be 0 for change points and 0.9 elsewhere for the hybrid setting. From Figure~\ref{fig:SVeventchange}, we observe that \method is able to perfectly recover all events and change points.

\subsubsection{\textbf{Resampled Setting}}
In this setting, we use a constant continuity rate of 0 for all time steps, i.e. the graph is resampled from the generative process at each step. In real world graphs, majority of the edges might not persist over consecutive time steps but rather determined by the underlying community structure. For the generation parameters and change points' details, we use the same setup as the Hybrid Setting. Resampling from the SBM model can be considered as a node level permutation within each community. As discussed in Section~\ref{sec:LAD}, the eigenvalues of the Laplacian is node permutation invariant. Indeed, \method outperforms all baselines. In comparison, EdgeMonitoring selects specific pairs of nodes to track over time thus being susceptible to node permutation and resulting in reduced performance.

\subsection{Real-world Experiments}
Here, we evaluate the performance of \method on two real-world benchmark datasets, as well as an original dataset. For UCI Message and Senate co-sponsorship experiments, \method is able to achieve strong performance using only the top 6 eigenvalues. For Canadian bill voting network, we report the LAD performance with the top 338 eigenvalues where 338 is the number of seats in the House of Commons of Canada. More information regarding each dataset can be found in Appendix~\ref{app:dataprocessing}.

\subsubsection{\textbf{UCI Message}} 
%
In this dataset, we treat each day as an individual time point. There are 59,835 total messages sent across all time stamps and 20,296 unique messages. To ensure privacy protection, all individual identifiers such as usernames, email and IP addresses were removed thus we use the dataset description provided in~\cite{panzarasa2009patterns} to find significant events in the dataset. We select the short term window to be 7 days as suggested by \citet{panzarasa2009patterns}. The long term window is then selected to be 14 days or two weeks.  

\begin{figure}[t]
    \begin{center}
        \centerline{\includegraphics[width=0.7\columnwidth]{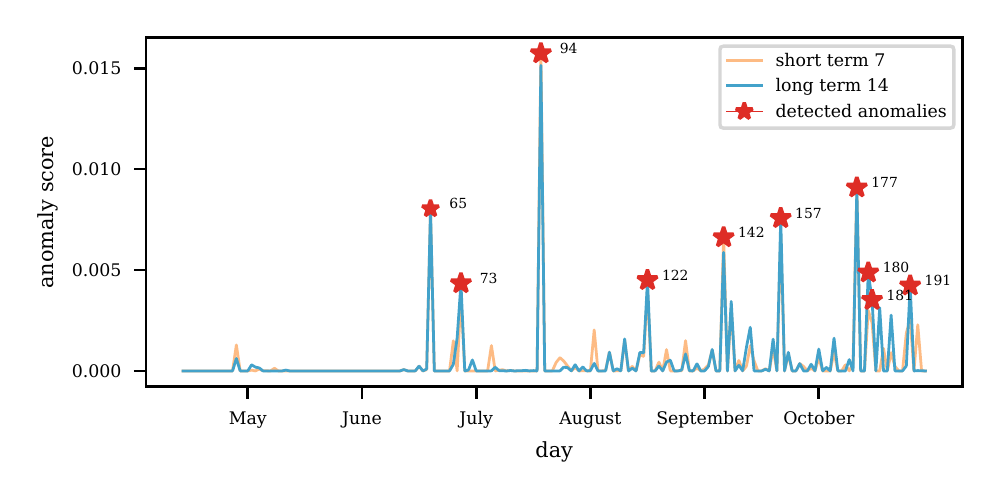}} 
        \caption{\method correctly detects the end of the university spring term and one day before the start of the fall term in the UCI message dataset.}
       \label{fig:UCIscore}
    \end{center}
    \vskip -0.3in
\end{figure}

Figure~\ref{fig:UCIscore} shows the $Z^*$ scores predicted from both the short term and long term window~(with top 6 eigenvalues). Panzarasa et al. mentioned that day 65\footnote{we set the index to start at 0} is the end of spring term and day 158 is the start of the fall term. Day 65 is correctly predicted by \method and activity vector while the top 10 predictions from EdgeMonitoring and TENSORSPLAT do not include either days. However, \method predicts day 157 as anomalous which corresponds to the day prior to the start of the fall term. It is likely that anomalous message behaviors are exhibited before school starts. As the edge weight~(number of characters in messages) shows important connections between users in social networks, the $Z$ scores has strongest correlation to the average edge weight per snapshot in this dataset.

\subsubsection{\textbf{Senate co-sponsorship Network}}

\begin{figure}[t]
    \begin{center}
        \centerline{\includegraphics[width=0.7\columnwidth]{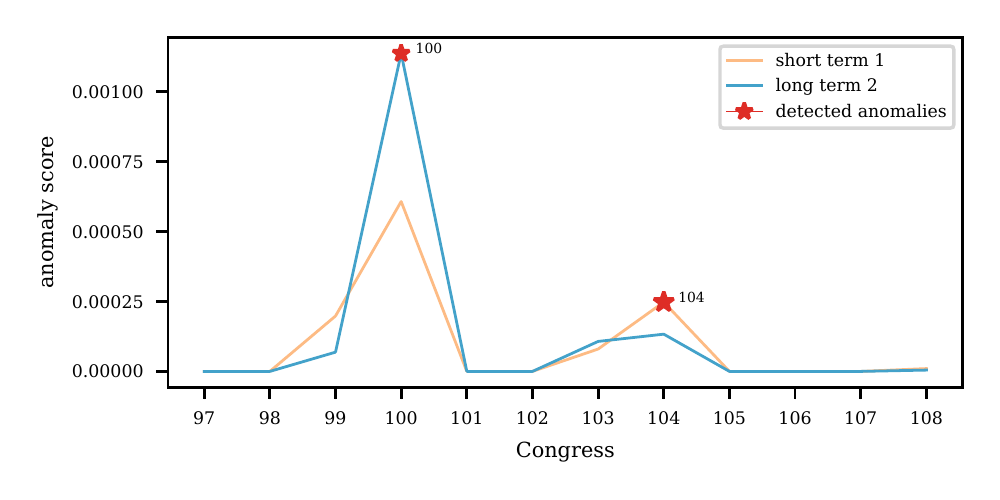}} 
        \caption{\method correctly detects the 100th and 104th congress as the top 2 most anomalous points.}
       \label{fig:senateScore}
    \end{center}
    \vskip -0.3in
\end{figure}

In this dataset, an edge is formed between two congresspersons if they are cosponsors on a bill. Fowler~\cite{fowler2006legislative} pointed out that the 104th Congress corresponds to ``a Republican Revolution'' which ``caused a dramatic change in the partisan and seniority compositions.'' Wang et al. also stated that the 104th Congress has the lowest clustering coefficient thus can be seen as a low point in collaboration while the 100th Congress has the highest clustering coefficient which signals significant collaboration. 

From Figure~\ref{fig:senateScore}, it is clear that \method can easily detect the above change points~(using only the top 6 singular values). Variations of \method that only utilizes the short term or the long term window are able to identify both points too. Wang et al. mentioned that EdgeMonitoring~\cite{wang2017fast} and LetoChange \cite{peel2015detecting} are able to detect both change points while DeltaCon~\cite{koutra2016deltacon} only predicts the 104th Congress.

The activity vector method requires full SVD which are computationally expensive and it is only able to detect the 100th Congress. The anomaly scores output by the activity vector method have similar magnitude thus making it difficult to identify change points. However, if we augment the activity vector method to use \method pipeline and aggregates two sliding windows, it is able to correctly predict both change points. This shows that aggregating the output from two sliding windows can improve empirical performance with a different graph embedding technique.

\subsubsection{\textbf{Canadian bill voting network}}

\begin{figure}[t]
    \begin{center}        \centerline{\includegraphics[width=0.7\columnwidth]{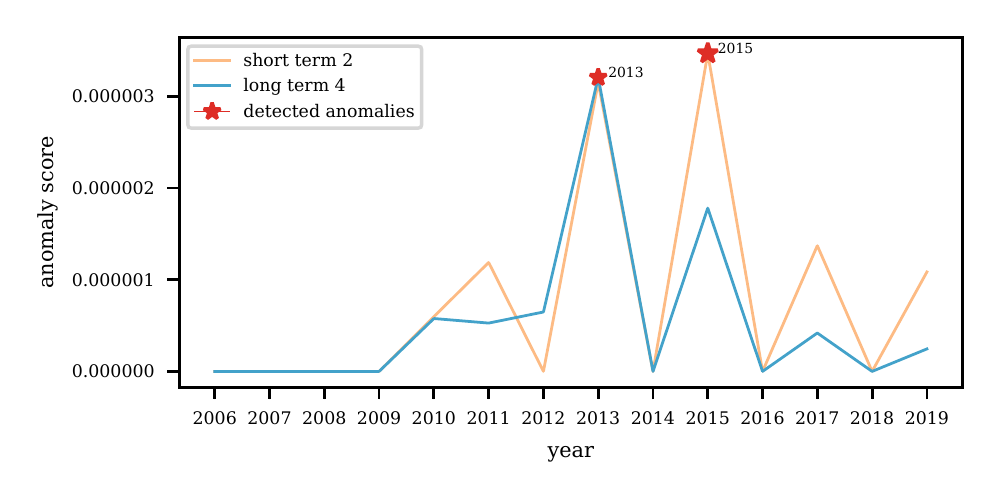}} 
    \caption{\method  predicts year 2013 and 2015 as anomalous years for the Canadian bill voting network.}
    \label{fig:canVotescore}
    \end{center}
    \vskip -0.2in
\end{figure}

To understand the temporal change in Canadian Parliament environment, we extracted open information\footnote{extracted from \url{https://openparliament.ca/}} to form the Canadian Parliament bill voting network. The Canadian Parliament consists of 338 Members of Parliament~(MPs), each representing an electoral district, who are elected for four years and can be re-elected. In the included time frame from 2006 to 2019, the increase in the number of electoral seats resulted in parliaments with different amounts of MPs; since 2015 the House of Commons has grown from 308 seats to 338. Naturally, the number of nodes in the network fluctuates from year to year. 2 year and 4 years are used as duration for the short term and long term windows respectively. We consider the MPs that voted yes for a bill to have a positive relation with the sponsor. In this way, a directed edge is then formed from a voter MP $u$ to the sponsor MP $v$ in a given graph snapshot. Each edge is then weighted by the number of times that $u$ voted positively for $v$.

\method detects 2013 and 2015 as two significant change points. Figure~\ref{fig:canVotescore} shows the $Z^*$ scores predicted by \method. 2013 is often considered as the year where cross party cohesion against Conservatives begins to decline. Prior to 2013, the Liberal and New Democratic Parties had more unity in voting patterns. As a new party leader~(Justin Trudeau) is elected in the Liberal party in 2013, changes in voting patterns are observed~\cite{marland2013political}. More details regarding data mining are discussed in Appendix~\ref{app:canadian}. In 2015, the house of common increased the number of constituencies from 308 to 338. Equally, on October 19th 2015 an election took place and the Liberal party won an additional 148 seats, with a total of 184 seats forming a majority government led by Justin Trudeau. Prior to 2015, the Liberal party was divided and however during this election things changed and literature shows the unified campaign at a local level from different members of parliament across the country~\cite{cross2016importance}.

\subsection{Summary of \method Results}

    \begin{table}[t]
        \resizebox{0.7\columnwidth}{!}{%
        \begin{tabular}{c | c c c | c | c}
        \toprule
        Metric & \multicolumn{3}{c|}{Hits @ 7} & Hits @ 10 & Hits @ 2\\
        \midrule
        Dataset & Pure & Hybrid & Resample & UCI & Senate \\
        \midrule
        \rowcolor{gray!20}LAD~(ours) & \textbf{100\%} & \textbf{100\%} & \textbf{100\%} & \textbf{50\%} & \textbf{100\%} \\
        
        EdgeMonitoring~\cite{wang2017fast} & \textbf{100\%} & \textbf{100\%} & 0\% & 0\% & \textbf{100\%} \\
        Activity vector~\cite{ide2004eigenspace} & 71.4\% & 0\% & 0\% & \textbf{50\%} & 50\% \\
        TENSORSPLAT~\cite{koutra2012tensorsplat} & 28.5\%& 14.2\% & 57.1\% & 0\% & 50\% \\
        \bottomrule
        \end{tabular}
        }
        \caption{\method  consistently finds significant anomalies across different datasets and outperforms alternative approaches. The hybrid experiment refers to the event and change point detection experiment.}
        \label{tab:LADresults}
        \vskip -0.2in
    \end{table}

Table~\ref{tab:LADresults} summarizes the empirical performances of \method and its comparison to alternative methods. We observe that \method has the best performance across all datasets. In this section, we discuss several experimental observations and provide  intuitions on the experimental results.  

In the UCI message and senate co-sponsorship experiments, \method achieves strong performance using only the top 6 singular values. It demonstrates that in real world datasets, a low rank truncated SVD is often sufficient to capture rich graph information. Together with efficient SVD computation techniques, \method can be scaled to large networks. In practice, the rank of the truncated SVD can be determined by the available computational resources. 

Lastly, synthetic experiment results suggest that \method can be used in a hybrid environment where both change points or events can occur. Therefore, \method is suitable to detect anomalous time points where it is possible for both one time events or fundamental changes in the network generative process to occur. It is often difficult to know beforehand what types of anomaly would appear in a dynamic network, \method would provide an effective and efficient solution without assumptions on the anomaly type. 





\definecolor{lightblue}{rgb}{0.9,0.98,1}
\definecolor{lightgreen}{rgb}{.9,1,0.95}

\section{Multi-view Experiments}\label{sec:multiview}
\subsection{Synthetic Experiments}

\renewcommand{\modtextbf}[1]{\noindent{\textbf{#1:}~}}

\changed{
In this section, we empirically examine the change point detection problem for multi-view dynamic graphs and evaluate our proposed \newmethod approach. In the synthetic experiments, we focus on answering the following questions: \modtextbf{Q1. Performance Improvement} How accurately can \newmethod detect synthetic anomalies by utilizing information from additional views as compared to state-of-the-art single view baselines and naive multi-view baselines? \modtextbf{Q2. Noise Invariant} How robust is \newmethod to noise  at individual views and can it leverage multi-view data to reduce the effect of noise? \modtextbf{Q3. Benefit of More Views} How will the performance of \newmethod change as we increase the number of incorporated views? and how the performance gain compares to the baselines?
}
\subsection{Baselines}
\changed{
Each data point in Figure~\ref{Fig:limit.and.noise}, \ref{Fig:moreviews} and Table~\ref{tab:multiResults} is the average over 30 trials and the standard deviation is reflected as the shaded area with the same color. All experiments are run on a desktop with AMD Ryzen 5 1600 CPU and 16GB installed RAM. Each \newmethod trial takes less than 5 minutes. We compare \newmethod with the following methods:}

\changed{
\modtextbf{LAD} We report \textit{the best} average individual view \method performance over multiple trials using the spectrum of unnormalized and normalized Laplacian~($\mat{L}$ and $\mat{L}_{sym}$) for comparison~(denoted by LAD and NL LAD, respectively). Note that in practice, this is overestimating the performance of such baseline as determining which view is most important is a difficult task.}

\changed{
\modtextbf{Naive Aggregation} We consider two naive aggregation strategies to extend \method into the multi-view setting. First, the maximum of the LAD anomaly score from each view~(per time step) is used as the aggregated anomaly score, named maxLAD. Second, the mean of the LAD anomaly scores is used instead and called meanLAD. Lastly, we apply the same naive aggregation approach to LAD scores obtained from the normalized Laplacian singular values and report them as ``NL maxLAD'' and ``NL meanLAD'' respectively.}

\changed{
\modtextbf{TENSORSPLAT} A multi-view dynamic graph can be represented as a 4th order tensor $\ten{T} \in \mathbb{R}^{T \times M \times N \times N}$ where $T, M, N$ are the total number of time steps, views and nodes respectively. After a rank $R$ CP decomposition of $\ten{T}$, one can obtain a signature vector of size $R$ for each time step $t$. To make this baseline more competitive, we then apply the same sliding window techniques used in LAD to identify the anomalies. Due to its high computational cost, we only report TENSORSPLAT~\cite{koutra2012tensorsplat} in Section~\ref{exp:limit} and average over 10 trials.}


\subsection{Performance Evaluation}
\changed{
Hits at $n$~($Hits@n$) metric reports the number of correctly identified anomalies out of the top $n$ time steps with the highest anomaly score. For synthetic experiment, we use the ground truth change points or events for evaluation. We extensively examine the performance of \newmethod as compared to baselines on two widely used graph generative model: the Stochastic Block Model~(SBM)~\cite{holland1983stochastic} and the Barabási-Albert (BA)~\cite{barabasi1999emergence} model. The accuracy reported in plots are the $Hits@7$ score for the 7 planted anomalies. The number of views are set to be 3 except for Section~\ref{exp:moreviews} and~\ref{exp:BA}. At each time step, the snapshot is re-sampled from its generative model. For all experiments, the sliding windows sizes are $w_s=5$ and $w_l=10$ respectively and the startup period~($0,\hdots,w_l$) is set to have an anomaly score of 0. For experiments involving the SBM model, change points correspond to splitting and merging communities and events are sudden increase in cross community connectivity.
}

\subsection{Case 1: SBM with increasing difficulty}~\label{exp:limit}
\changed{
To answer if \newmethod leverages multi-view information to improve change point detection, we examine a sparse SBM model with an increasing level of difficulty. In a SBM model with $n$ nodes, $p_{in} = \frac{c_{in}}{n}$ and $p_{ex} = \frac{c_{out}}{n}$ are the edge probabilities for the within community and cross community edges. The constants $c_{in}$ and $c_{out}$ are used to adjust the strength of community connectivity. 
In this experiment, we set $c_{in} = 12$ and consider different values for $c_{out}$; note that as $c_{in} - c_{out}$ gets smaller, the task increases in difficulty. As events would be easy to detect when the difference between $c_{in}$ and $c_{out}$ is small, we only consider change points for this experiment~(see Table~\ref{tab:SBM} (a)). 
}

\begin{table*}[t]
\caption{Experiment Setting: the changes in the generative model in \\
(a) Section~\ref{exp:limit} and \ref{exp:moreviews} where the SBM parameter $c_{in} = 12, p_{in} = 0.024$ and $ p_{ex} = \frac{c_{out}}{\# nodes}$. The more communities there are, the higher the color intensity is. \\
(b) Section~\ref{exp:noisy} where the SBM parameter $p_{in} = 0.024, p_{ex}=0.004$ and both events and change points are observed.} \label{tab:SBM}
\begin{centering}
\begin{subtable}[t]{0.4\columnwidth}
\begin{tabular}{c | l | c}
\toprule
\multicolumn{3}{c}{\modtextbf{Change Points} SBM Model parameters}\\
\midrule
Time Point & Type & $N_c$\\
\midrule
0 & start point &  2\\
\rowcolor{blue!4} 16 & change point & 4\\
\rowcolor{blue!6} 31 & change point & 6\\
\rowcolor{blue!10} 61 & change point & 10\\
\rowcolor{blue!20} 76 & change point & 20\\
\rowcolor{blue!10} 91 & change point & 10\\
\rowcolor{blue!6} 106 & change point & 6\\
\rowcolor{blue!4} 136 & change point & 4\\
\bottomrule
\end{tabular}
\caption{anomalies in Section~\ref{exp:limit} and~\ref{exp:moreviews}}
\end{subtable}%
\begin{subtable}[t]{0.5\columnwidth}
\begin{tabular}{c | l | c | c | c}
\toprule
\multicolumn{5}{c}{\modtextbf{Events \& Change Points} SBM Model parameters}\\
\midrule
Time Point & Type &  $N_c$ & $ p_{in}$ & $ p_{ex}$\\
\midrule
0 & start point & 4 &  0.024 &  0.004 \\\rowcolor{lightgreen}
16 & event & 4 &  0.024 & \textbf{0.012} \\\rowcolor{blue!5}
31 & change point & \textbf{10} &  0.024 &  0.004 \\\rowcolor{lightgreen}
61 & event & 10 &  0.024 & \textbf{0.012} \\\rowcolor{blue!5}
76 & change point & \textbf{2} &  0.024 &  0.004 \\\rowcolor{lightgreen}
91 & event & 2 &  0.024 & \textbf{0.012} \\\rowcolor{blue!5}
106 & change point & \textbf{4} &  0.024 &  0.004 \\\rowcolor{lightgreen}
136 & event & 4 &  0.024 & \textbf{0.012} \\
\bottomrule
\end{tabular}
\caption{anomalies in Section~\ref{exp:noisy}}
\end{subtable}%
\end{centering}
\end{table*}

\begin{figure}[t]
\centering
\begin{subfigure}[b]{0.8\columnwidth}\centering
 \includegraphics[width=.8\columnwidth]{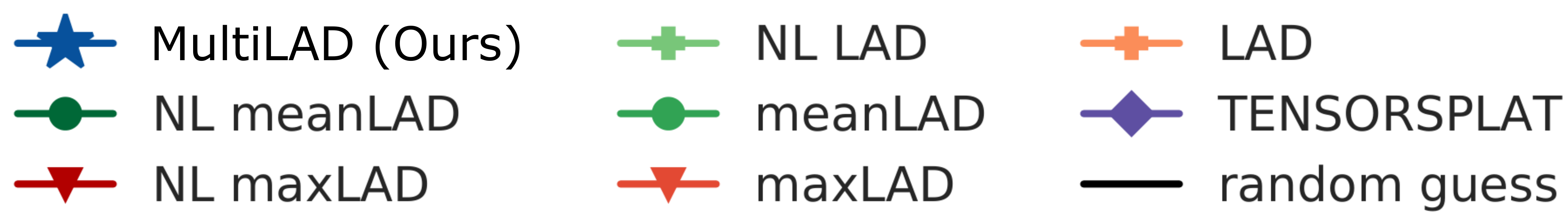}
\end{subfigure}%

\begin{subfigure}{0.4\columnwidth}
 \includegraphics[width=\columnwidth]{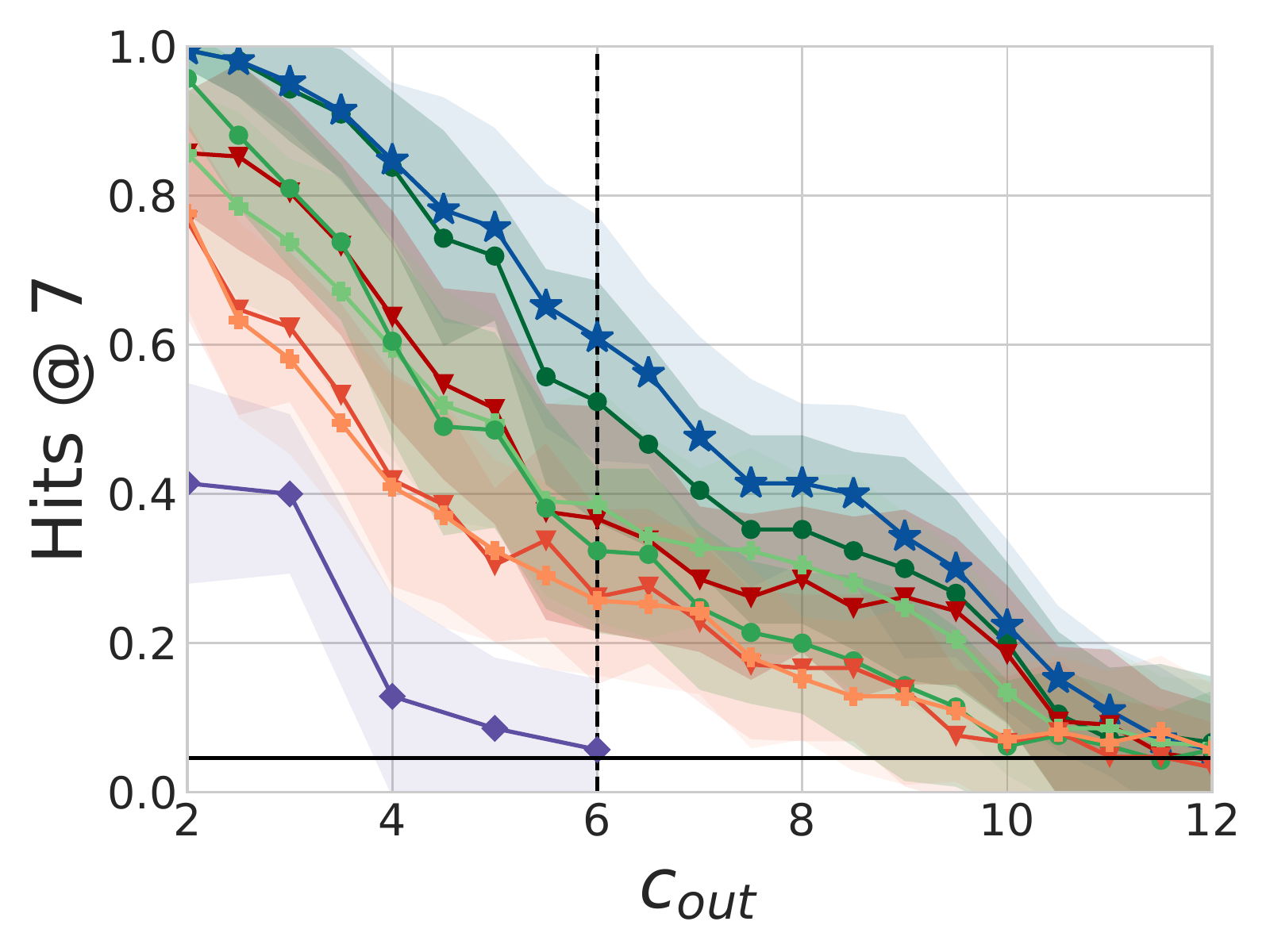}
  \caption{SBM no noise}
\end{subfigure}%
\begin{subfigure}{0.4\columnwidth}
  \centering
  \includegraphics[width=\columnwidth]{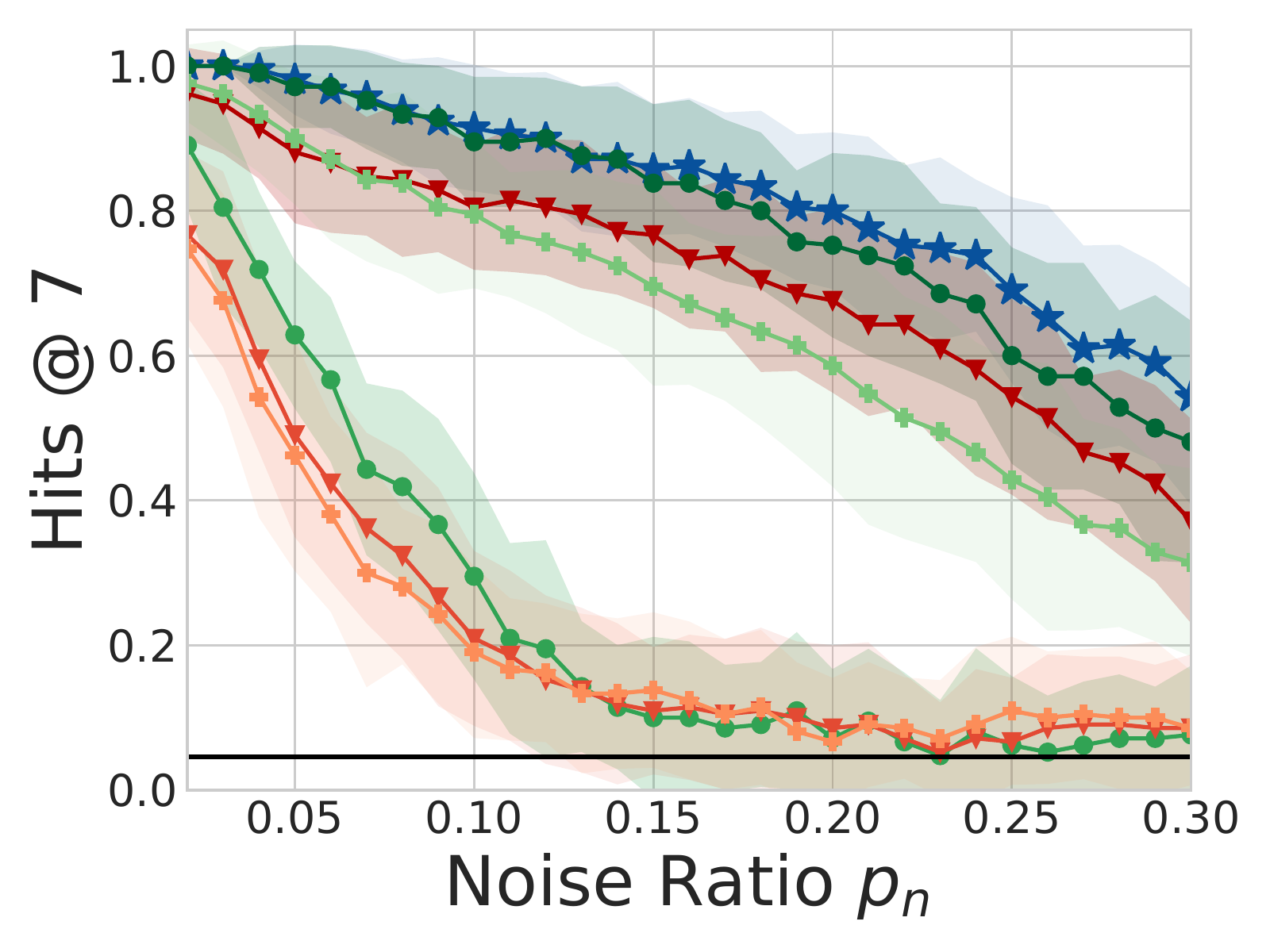}
  \caption{SBM with noise}
\end{subfigure}%

\caption{(a) \newmethod  significantly outperforms all baselines as the inter-community connectivity $c_{out}$ approaches the intra-community connectivity $c_{in}=12$. We also indicate the community detection limit for spectral clustering methods with a dotted vertical line. 
\\ (b) \newmethod effectively leverages multi-view information to filter out noise injected into individual views while single view methods such as LAD is easily disrupted by high amount of noise.}
\label{Fig:limit.and.noise}
\vskip -0.2in
\end{figure}

\changed{
Figure~\ref{Fig:limit.and.noise}~(a) shows that \newmethod significantly outperforms all baselines as the community structure becomes harder to detect. \newmethod is able to aggregate over 3 views and significantly outperform the single view method LAD by at as much as $48.3\%$ at $c_{out}=4$. \newmethod also outperforms the best naive multi-view baseline NL meanLAD by $9.5\%$ at $c_{out}=7$. Note that for $c_{out} > 11$, the community structure is almost non-existent as the graph closely resembles an Erdős-Rényi graph~\cite{erdHos1960evolution} where all edges are sampled with equal probability. As a reference for the strength of community structure, we indicate the community detection limit for spectral clustering methods on two equal sized communities as a dotted vertical line in the figure. The limit is reached when $c_{in} - c_{out} = \sqrt{2(c_{in} + c_{out})}$~\cite{newman2016structure} and in this case, $c_{in} = 12$ and $c_{out}=6$. We observe that after $c_{out}=6$, \newmethod still detects the change points to a significant extent while outperforming other methods. Although being a multi-view baseline, TENSORSPLAT's performance is much worse than \newmethod and even single-view LAD. Designing an effective multi-view method is an important direction.
}

\subsection{Case 2: all views are perturbed by noise}
\label{exp:noisy}

\changed{
To examine if \newmethod is robust to noise from individual views, we add noise to the SBM generative process at each view by examining each pair node pair $i,j$ and flip the entry $\mathcal{A}_{i,j}$ of the adjacency matrix to $1-\mathcal{A}_{i,j}$ with probability $p_n$. In this experiment, we set $p_{in}=0.024$ and $p_{ex}=0.004$ and adjust $p_n$ for the noise level. We also include both events and change points (see Table~\ref{tab:SBM} (b)).
}

\changed{
Figure~\ref{Fig:limit.and.noise}~(b) demonstrates that \newmethod is highly robust to noise and outperforms all baselines. Notably, single-view methods like LAD are highly susceptible to noise and their performance degrades quickly as higher ratios of noise are injected, In comparison, \newmethod maintains strong performance despite each view having a high level of noise. This shows that using the scalar power mean to aggregate singular values from individual views is a very effective way to filter out noise. When $p_n = 0.30$, \newmethod outperforms single-view LAD by $45.7\%$ and NL LAD by $22.9\%$.
\newmethod upper bounds all naive multi-view baselines and significantly outperforms NL maxLAD and NL meanLAD when the noise ratio is high~(i.e. $p_n > 0.15$).
}

\subsection{Case 3: SBM with additional views}
\label{exp:moreviews}

\begin{table}[t]
 \centering
\resizebox{0.4\columnwidth}{!}{%
\begin{tabular}{c | l | c}
\toprule
\multicolumn{3}{c}{\modtextbf{Change Points} BA Model parameters}\\
\midrule
Time Point & Type & $m$\\
\midrule
0 & start point &  1\\\rowcolor{red!2}
16 & change point & 2\\\rowcolor{red!4}
31 & change point & 3\\\rowcolor{red!7}
61 & change point & 4\\\rowcolor{red!10}
76 & change point & 5\\\rowcolor{red!13}
91 & change point & 6\\\rowcolor{red!17}
106 & change point & 7\\\rowcolor{red!20}
136 & change point & 8\\
\bottomrule
\end{tabular}
}
\caption{anomalies in Section~\ref{exp:BA} where the BA model parameter $m$ is the number of edges to attach from a new node to existing nodes. The increased color intensity in the table indicates higher $m$ value.} \label{tab:BA}
\end{table}

\begin{figure}[t]
\centering
\begin{subfigure}[b]{0.8\columnwidth}\centering
 \includegraphics[width=0.8\columnwidth]{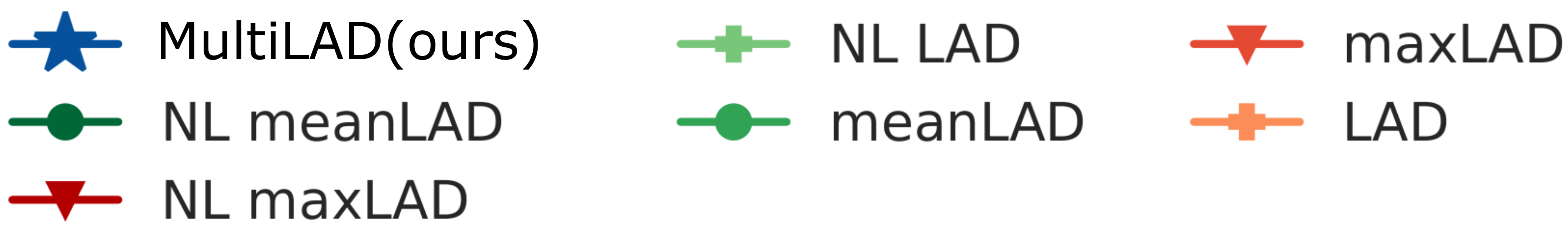}
\end{subfigure}%

\begin{subfigure}{0.4\columnwidth}
 \includegraphics[width=\columnwidth]{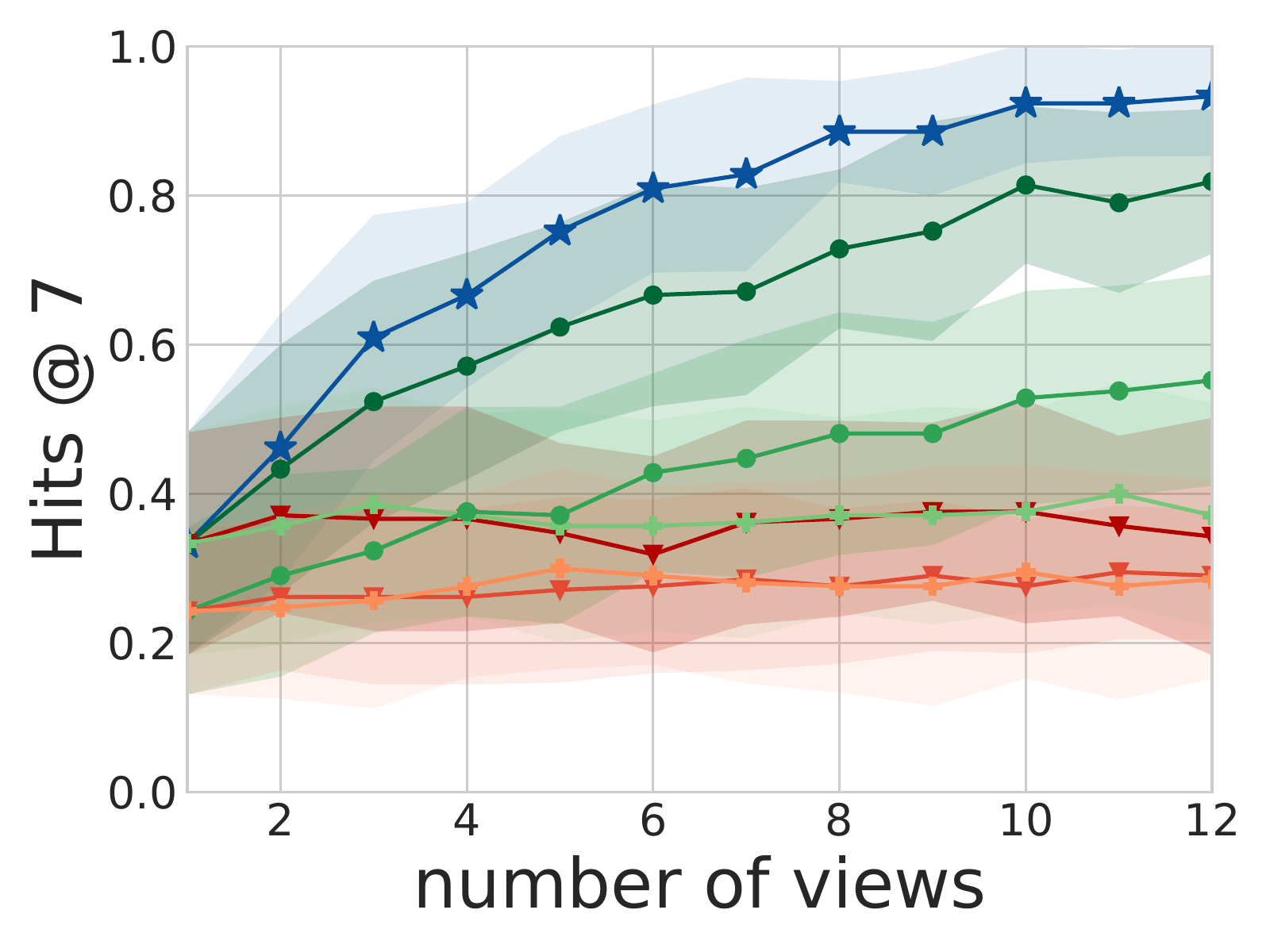}
  \caption{SBM}
\end{subfigure}%
\begin{subfigure}{0.4\columnwidth}
  \centering
  \includegraphics[width=\columnwidth]{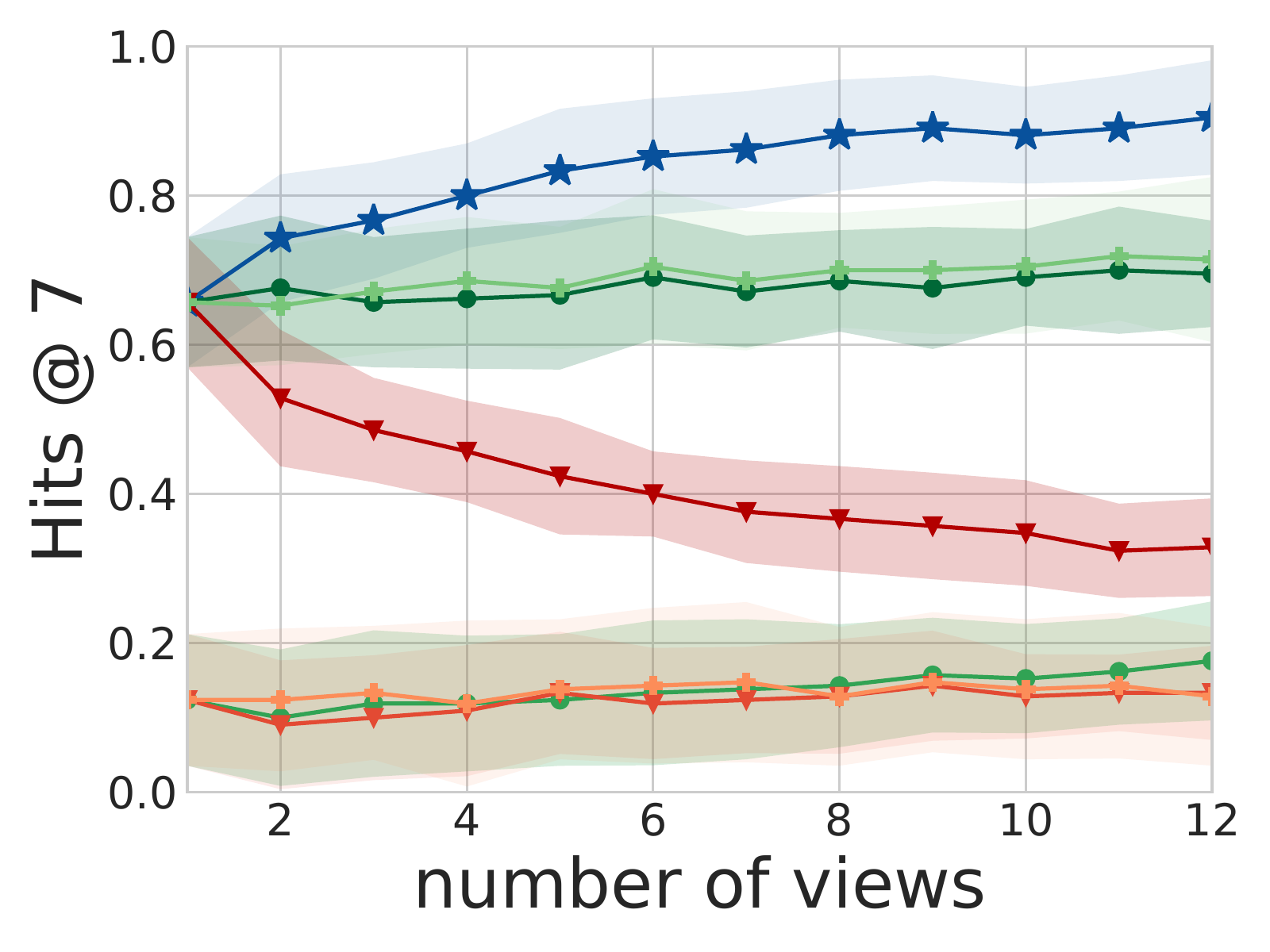}
  \caption{BA}
\end{subfigure}%

\caption{\newmethod sees the most performance gain from additional views in the SBM model setting~(a) and BA model setting~(b) when compared to all baselines.}
 \label{Fig:moreviews}
\vskip -0.2in
\end{figure}

\changed{
To understand how \newmethod benefits from having additional views, we follow the same setting as in Section~\ref{exp:limit} and add more views. For this experiment, we fix $p_{in}=0.024$ and $p_{ex}=0.012$, and compare the performances of all methods when the number of available views increases.
}

\changed{
Figure~\ref{Fig:moreviews}~(a) shows that \newmethod benefits the most from having additional views. Single view methods such as LAD are unable to take advantage of the multi-view data thus remaining at the same performance. Another interesting observation is that max aggregation barely benefits from the additional views as it can be easily dominated by the anomaly score from one particular view. Both NL meanLAD and meanLAD improve with additional views as they compute the average of the anomaly score thus converging to the true expected anomaly score. 
The strong performance of \newmethod can be attributed to the fact that \newmethod directly aggregates the singular values while NL meanLAD and NL maxLAD are dependent on the anomaly score output from individual views. The biggest performance gap appears at 7 views where \newmethod outperforms NL maxLAD by $46.7\%$ and NL meanLAD by $15.7\%$.
}

\subsection{Case 4: all views are BA models}~\label{exp:BA}


\changed{
Lastly, we further investigate \newmethod performance in the Barabási-Albert~(BA) model to show that \newmethod is effective beyond the SBM model. In this experiment, the change points correspond to the densification of the network~(parameter $m$, increased number of edges attached from a new node to an existing node). The details are described in Table~\ref{tab:BA} (c). 
}

\changed{
Figure~\ref{Fig:moreviews}~(b) shows that \newmethod benefits significantly from more views in the BA setting as well. In particular, with just 2 views, \newmethod outperforms LAD by 61.9\%, NL LAD by 9\% and NL meanLAD by 6.7\%. \newmethod is able to efficiently utilize multi-view information as soon as additional views are available and benefits even more when there are more views. We also observe that NL maxLAD has decreasing performance when the number of views increased. This is likely due to naively aggregating the max anomaly score from all views and picking up the noise from additional views rather than useful information. This shows that naive aggregation strategies can have adverse effects. Also, NL meanLAD performs similarly to single view NL LAD thus is not able to effectively leverage the multi-view information.
}

\subsection{Summary of \newmethod Results}

\begin{table}[t]
    \caption{\newmethod outperforms all alternative approaches on different datasets.}
    \centering
    \resizebox{0.7\columnwidth}{!}{%
    \begin{tabular}{l| r r r r}
    \toprule
    Metric & \multicolumn{4}{c}{Hits @ 7} \\
    \midrule
    Experiment                                  & SBM~(3v)                     & SBM~(noisy)             & SBM~(12v) & BA~(12v) \\
    \midrule
    \rowcolor{gray!20}\newmethod                   & \textbf{.61 $\pm$ .16} & \textbf{.86 $\pm$ .09} & \textbf{.94 $\pm$ .08} & \textbf{.90 $\pm$ .08}\\
    LAD~\cite{huang2020laplacian}        & $.26 \pm .10$          & $.14 \pm .11$         & $.29 \pm .13$ & $.13 \pm .09$\\
    maxLAD                                      & $.26 \pm .12$          & $.11 \pm .09$          & $.29 \pm .09$ & $.13 \pm .06$\\
    meanLAD                                     & $.32 \pm .11$          & $.10 \pm .11$         & $.55 \pm .14$ & $.18 \pm .08$\\
    NL LAD                                      & $.39 \pm .16$          & $.70 \pm .14$         & $.37 \pm .15$ & $.71 \pm .11$\\
    NL maxLAD                                   & $.37 \pm .15$          & $.77 \pm .10$         & $.34 \pm .16$ & $.33 \pm .07$\\
    NL meanLAD                                  & $.52 \pm .16$          & $.84 \pm .11$         & $.82 \pm .10$ & $.70 \pm .07$\\
    TENSORSPLAT~\cite{koutra2012tensorsplat}   & .$6 \pm .9\ $            & N/A                     & N/A & N/A \\
    \bottomrule
    \end{tabular}}
    \label{tab:multiResults}
\end{table}

\changed{
Table~\ref{tab:multiResults} summarizes the performance comparison between \newmethod and other baselines across different experiments. We use k`v' to indicate the number of views in the network. For SBM~(3v) and SBM~(12v), we report the setting where $p_{in} = 0.024$ and $p_{ex}=0.012$. For noisy SBM setting, we follow Section~\ref{exp:noisy} and set noise to be 0.15. For BA, we follow the set up in Section~\ref{exp:BA}.
}

\changed{
In all settings, \newmethod significantly outperforms all baselines. By efficient use of multi-view information, \newmethod improves upon LAD by as high as 62\% on the SBM experiments and 77\% on the BA experiment. \newmethod also sees increased gain as additional views become available, achieving 33\% gain when using 12 views instead of 3. This shows that \newmethod effectively leverages the additional views to achieve stronger performance.
}

\changed{
\newmethod outperforms alternative aggregation strategies including max and mean across all settings and is more robust to noise which shows the benefit of merging information at the signature vector level rather than the anomaly score level. Moreover, max and mean baselines exhibit inconsistent behaviors on different settings. On the BA model, the mean operation does not benefit from having additional views while it is able to outperform single-view baselines on the SBM model. The max operation is detrimental to performance with additional views on the BA model, but not on the SBM model. In contrast, \newmethod always benefits from additional views.
}
\changed{
We can also see that it is better to use the normalized Laplacian (NL LAD) over the unnormalized one (LAD). It is known that on large dense graphs, the distribution of the eigenvalues of the Laplacian is close to the distribution of the degrees of the vertices~\cite{vizuete2020laplacian, hata2017localization}. After normalization by node degree, the singular values of the normalized Laplacian can better reflect the graph structure, thus be more suited for change point detection. 
}

\subsection{Real World Experiments} \label{Sec:real}

\begin{figure}[t]
\centering

\begin{subfigure}{0.5\columnwidth}
 \includegraphics[width=\columnwidth,trim={0in 0in 0 0 },clip]{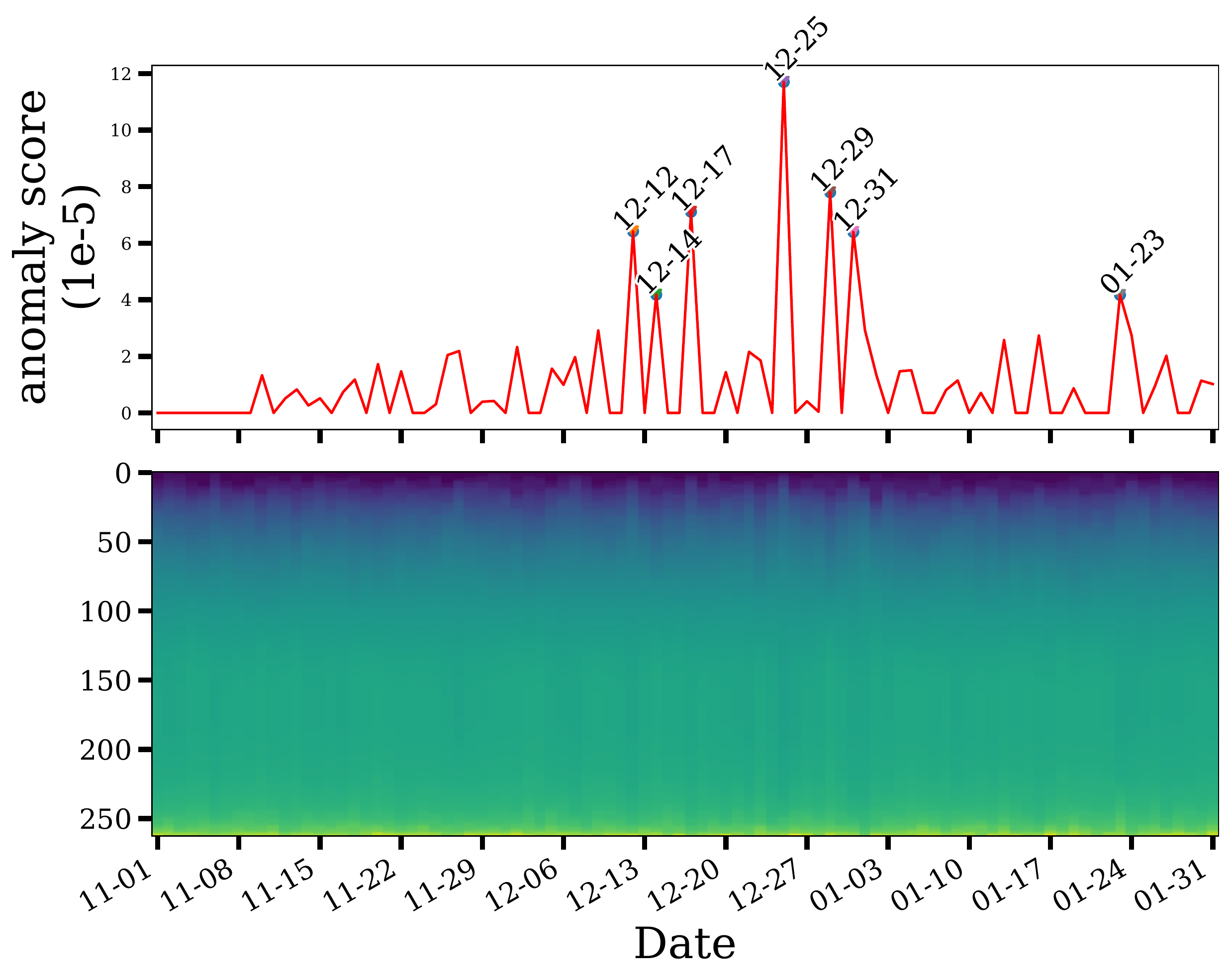}
  \caption{NYC TLC 2015-2016}
  \label{Fig:NYCTLC2015}
\end{subfigure}%
\begin{subfigure}{0.5\columnwidth}
  \centering
  \includegraphics[width=\columnwidth,trim={0in 0in 0 0 },clip]{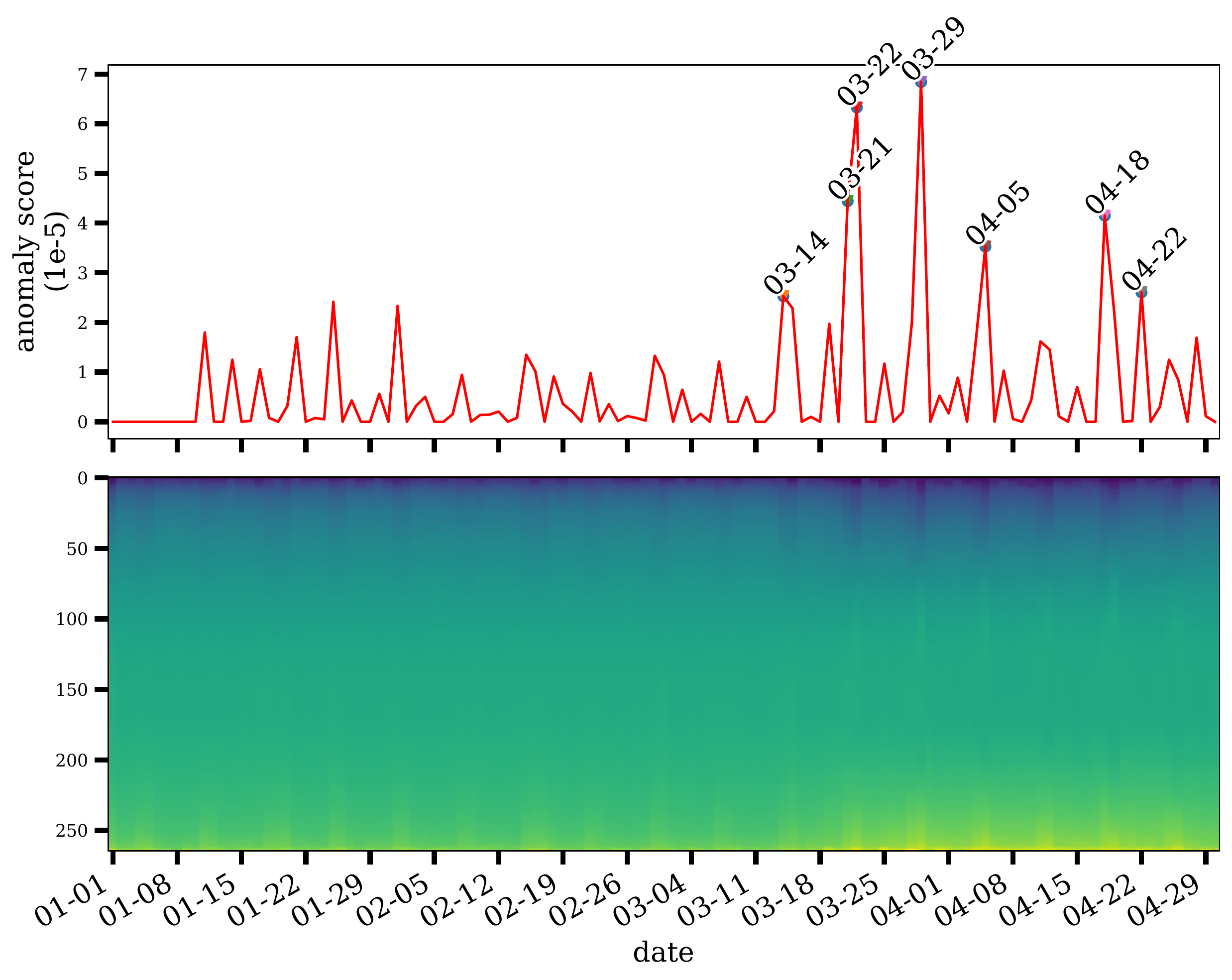}
  \caption{NYC TLC 2020}
\end{subfigure}%

\caption{Anomaly score (top, annotated with the 7 most anomalous time points) and \newmethod's aggregated spectrum~(bottom) for~(a). the NYC TLC 2015-2016 dataset and~(b).NYC TLC 2020 dataset.}
 \label{Fig:NYCTLC}
\vskip -0.2in
\end{figure}


\changed{
Next, we examine \newmethod in the real world setting.
We used New York City~(NYC) Taxi \& Limousine Commission (NYC TLC)\footnote{\url{https://www1.nyc.gov/site/tlc/about/tlc-trip-record-data.page}} trip records to construct and analyze two dynamic multi-view graphs at different time periods. One of them is a novel multi-view dynamic graph dataset set in 2020, showcasing \newmethod's ability to detect key dates in the implementation of New York City's COVID-19 response. Details on data processing are provided in Appendix~\ref{app:dataprocessing}. We use window sizes of 3 and 7 days in our experiments to model both short-term events and long-term change points.
}

\changed{
\modtextbf{NYC TLC 2015-2016} From NYC TLC records, we acquired the timestamped trip data from two taxi companies (green and yellow) covering November 2015 to January 2016 similar to~\cite{SPOTLIGHT}. We used these records to construct a dynamic two-view directed weighted graph over 264 nodes~(representing taxi zones) in which each time point represents a distinct day.
}

\changed{
Figure~\ref{Fig:NYCTLC} shows \newmethod's anomaly score curve and the corresponding Laplacian spectrum (aggregated using the scalar power mean  approach). In the spectrum visualization, the time points with high anomaly scores are visually distinct from their surrounding points thus confirming that \newmethod's aggregated spectrum correctly retains meaningful information from individual views. Comparing \newmethod's results to those reported by~\cite{SPOTLIGHT} revealed that there are many high-scoring anomalies in common~(including Christmas day, New Year's Eve, New Year's Day, and the blizzard on January 23\textsuperscript{rd}.) There are some minor differences in the anomalies identified by \newmethod and SPOTLIGHT which can be attributed to the fact that \newmethod is analyzing the multi-view version of this dataset while SPOTLIGHT is operating on the single-view version~(these differences are discussed further in Appendix~\ref{app:NYC2015}).
}


\changed{
\modtextbf{NYC TLC 2020} We also extracted NYC TLC data from January 1st 2020 to April 30th 2020. There are 4 views in total with 2 additional views corresponding to for-hire vehicle (FHV; pre-arranged transportation services like community cars and limousines\footnote{\url{https://www1.nyc.gov/site/tlc/businesses/for-hire-vehicle-bases.page}}) and high-volume for-hire vehicle (FHVHV; Uber, Lyft, and Via\footnote{\url{https://www1.nyc.gov/site/tlc/businesses/high-volume-for-hire-services.page}}) transportation services were available for this timecourse.
}

\changed{
We hypothesized that anomalies in the 2020 NYC transit dataset would align with key time points in the rollout of NYC's COVID-19 response. Our results suggest this hypothesis was largely correct (see Figure~\ref{Fig:NYCTLC} and Appendix~\ref{app:NYC2020}): \newmethod is able to identify the adoption of the ``NYS on Pause Program'' (all non-essential workers must stay home, March 22\textsuperscript{nd}), March 29\textsuperscript{th} (the day after all non-essential construction sites were halted in NYS), and April 18\textsuperscript{th} (the Saturday after the first extension of the ``NYS on Pause Program'', April 16\textsuperscript{th}). These results show that \newmethod can indeed detect significant and meaningful real world events from multiple views. In this case, considering the 4 views individually and determining the most informative one is unpractical.
}

\section{Conclusion}

\changed{
In this work, we introduced two novel change point detection method : \method and \newmethod for single-view and multi-view dynamic graphs respectively. \method utilizes the singular values of the Laplacian matrix to embed graph snapshots and explicitly captures both the short term and the long term temporal relations to detect events and change points. Through evaluations on both synthetic and real world datasets, we showed that \method is more effective at identifying significant events than previous state-of-the-art methods. Next, on multi-view dynamic graphs, \newmethod aggregates the singular values of the normalized Laplacian matrices through the scalar power mean operation and identifies the most informative singular values from each view. On synthetic benchmarks, \newmethod benefits from additional views, is robust to noise, and outperforms competitive single view and multi-view baselines. On real world traffic networks, we showed that \newmethod correctly identifies significant events which drastically disrupts the flow of traffic. 
}
\section{Acknowledgments}
This research was supported by the Canadian Institute for Advanced Research (CIFAR AI chair program), Natural Sciences and Engineering Research Council of Canada (NSERC) Postgraduate Scholarship-Doctoral (PGS D) Award and Fonds de recherche du Québec – Nature et Technologies (FRQNT) Doctoral Award.





\bibliographystyle{ACM-Reference-Format}
\bibliography{main}

\newpage
\appendix


\section{Additional Materials for \method Experiments} \label{app:repro}
This section contains additional material for \method and single-view experiments in Section~\ref{SV:dataset}. We report the implementations used in our experiments. 
For TENSORSPLAT, we first compute the PARAFAC decomposition~(using Tensorly~\cite{kossaifi2019tensorly} library in python) to obtain the temporal factors. Then the scikit-learn~\cite{scikit-learn} python implementation of the Local Outlier Factor algorithm. For EdgeMonitoring method, we use the matlab code kindly provided by the original authors and keep the default parameters.

\subsection{The effect of $Z^*$ scores} \label{app:zscore}

\begin{figure}[t]
    \begin{center}
        \centerline{\includegraphics[width=0.8\columnwidth,trim={0in 0in 0 0 },clip]{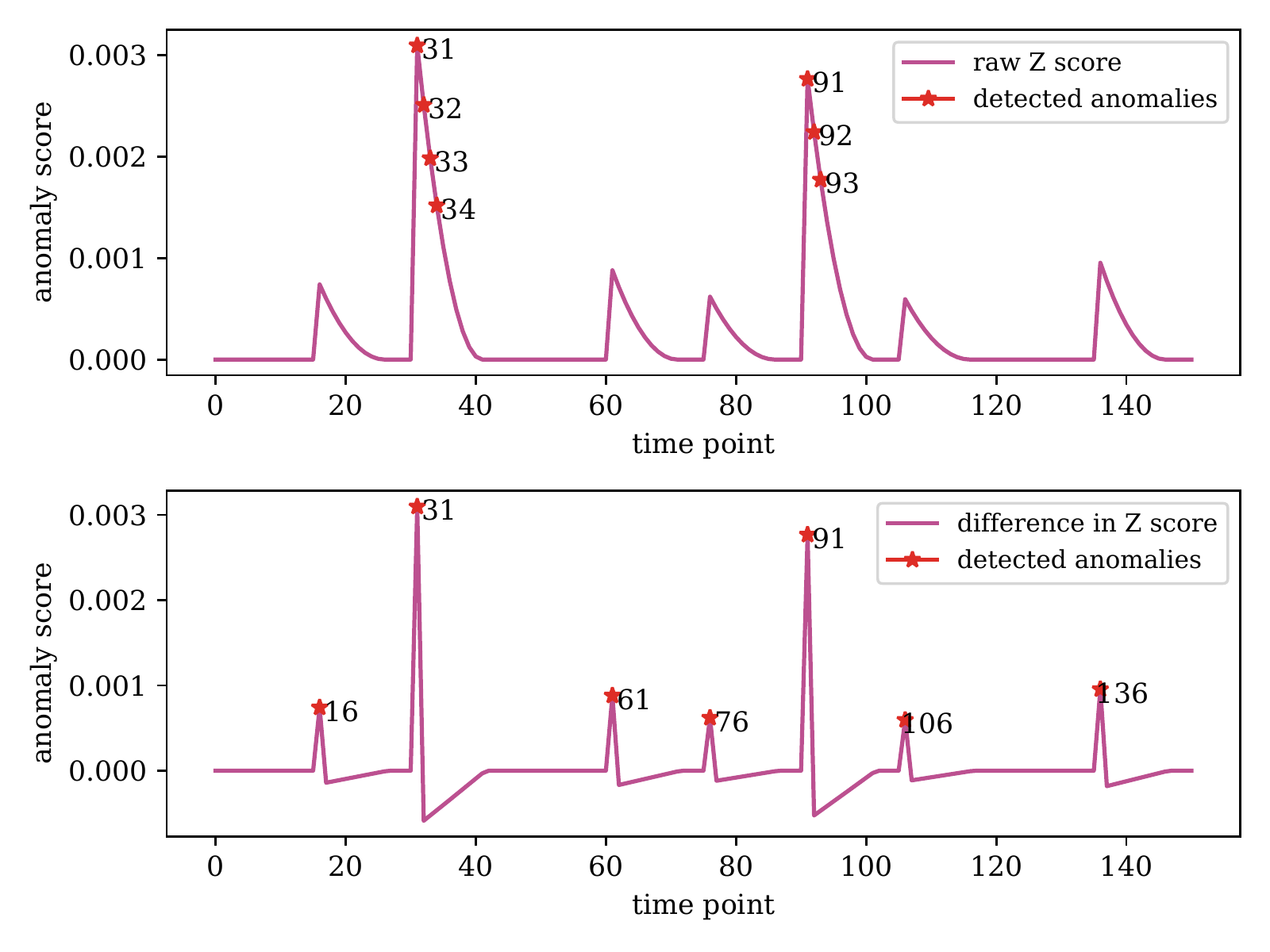}} \vskip -0.2in
        \caption{\method perfectly recovers the injected change points of Table~\ref{tab:SVSBM}. We visualize the predicted anomaly scores. Both the raw $Z$ score (top) and the difference in consecutive $Z$ scores, the $Z^*$ scores (bottom) are plotted.}
       \label{fig:SVSBMscore}
    \end{center}
\end{figure}

In Section~\ref{sec:LAD}, we defined the anomaly score as the difference in $Z$ score when compared to the previous time step. The final anomaly score is defined as $Z_t^* = min(Z_t - Z_{t-1},0)$. The points with the largest $Z^*$ are then selected as anomalies. Here we visualize both the raw $Z$ score and $Z^*$ score in Figure~\ref{fig:SVSBMscore} for the pure setting in Section~\ref{sec:pure}. Note that when using the raw $Z$ scores, we observe declines in anomaly score after the change point. In a sliding window which contains graph snapshots from the previous graph generative process and the current one, the normal behavior vector is computed based on a mix of graphs from two generative processes thus leading to the observed declining anomalous scores after the change point. In comparison, using $Z^*$ scores makes \method more robust under different choices of sliding windows.

\subsection{Spectral Properties and Their Connections} \label{app:spec}

\begin{table}[ht]
\centering
\begin{tabular}{c|c}
\hline
\hline
Spectral Properties & Connections \\
\hline 
$\mathbf{L},\mathbf{L}_{rw},\mathbf{L}_{sym}$ eigenvalues & connected components~\cite{von2007tutorial} \\

$\mathbf{L}$ eigenvectors & ratio cut~\cite{hagen1992new} \\

$\mathbf{L}_{rw},\mathbf{L}_{sym}$ eigenvectors & normalized cut~\cite{shi2000normalized,von2007tutorial} \\

$\mathbf{A}$ eigenvalues & KATZ centrality~\cite{goh2001universal}\\
$\mathbf{A}$ eigenvectors & eigenvector centrality~\cite{bonacich1987power}\\
$\mathbf{A}$ dominant eigenvector &stationary distrib., PageRank~\cite{page1999pagerank}\\
\hline
\end{tabular}
\caption{Spectral Properties and Their Connections}
\label{tab:connections}
\end{table}

The above table summarizes connections between different spectral properties in the graph and their connections in the literature. We use the same notation as~\cite{von2007tutorial}. $\mathbf{L},\mathbf{L}_{rw},\mathbf{L}_{sym}, \mathbf{A}$ represent the unnormalized Laplacian matrix, the random walk Laplacian matrix, the symmetric Laplacian matrix and the adjacency matrix respectively. 

\subsection{Interpreting Results}
It is often difficult to interpret the anomaly score of a given change point detection method as the task inherently demands direct comparison between global graph structures over time. As network characteristics vary drastically across domains~\cite{broido2019scale}, it is important to design metrics that help us understand the correlation between  anomaly scores and well-known graph properties. In this work,we identify temporal outliers in specific graph properties and compare them to the ones predicted by \method . We choose the outlier score $y$ as follows:
\begin{equation}
    y = \frac{|\alpha_t - \alpha_{avg} |}{\alpha_{std}},
\end{equation}
where $\alpha_{avg}$, $\alpha_{std}$ are the average and standard deviation of $\alpha$ computed from a moving window. We select the moving window size to correspond with the short term window size $s$. Then we compute the Spearman rank correlation~\cite{ziegel2001standard} to understand the statistical dependence between two ranked variables. We observe that \method is not relying on one particular graph statistics but rather on the most important aspects of the dynamic graph of interest. 

\begin{table}[t]
\centering
\begin{tabular}{l|c}
\hline
\hline
Graph Property & Spearman Rank Correlation \\
\hline 

\# of connected components &\textbf{ 17.0}\% \\
Transitivity & 11.8\% \\
\# of edges & 7.5\% \\
Average degree per node & 15.9\% \\
\hline
\end{tabular}
\caption{\method scores correlate well with the number of connected components when injected points are changing number of blocks in SBM (Table~\ref{tab:SVSBM}). Spearman rank correlation between \method and other graph properties is also reported.}
\label{tab:rank}
\end{table}

By examining Spearman rank correlations in Table~\ref{tab:rank}, we observe that \method predictions are most strongly correlated with the number of connected components while still having a positive correlations with other network properties such as transitivity. As \method captures high level graph structures, it is sensitive to the important properties in the dynamic graph of interest while not dependent on any single property.

We observe that \method is not dependent on any one particular graph statistics. Empirically, \method correlates with different graph properties depending on the network of interest. This coincide with our intuition that any type of significant change in the graph structure could disrupt the singular values. These experimental results suggest that \method can capture changes of different nature in the graph structure as it tracks the compression loss of the Laplacian matrix to low rank approximations. 

\subsection{UCI Message Network}

The UCI Message dataset is a directed and weighted network based on an online community of students at the University of California, Irvine. Each node represents a user and each edge encodes a message interaction from one user to another. The weight of each edge represents the number of characters sent in the message. When an user account is created, a self edge with unit weight is added. A total of 1,899 users was recorded. The network data covers the period from April to October 2004 and spans 196 days. 

\subsection{Senate co-sponsorship Network}

Senate co-sponsorship network~\cite{fowler2006legislative} examines social connections between legislators from their co-sponsorship relations on bills during the 93rd-108th Congress. Bills are grouped into temporal snapshots biannually~(time frame for each graph) and co-sponsors on a bill form a clique. Similar to~\cite{wang2017fast}, we start from the 97th Congress as full amendments data is available from there onward.

\subsection{Canadian bill voting network} \label{app:canadian}

The Canadian bill voting network was mined from the Open Parliament API~(http://api.openparliament.ca/), a source for digitized data from the House of Commons in JSON format. We first extracted all MPs in the canadian parliament from 2006 to 2019. The network nodes for each snapshot only includes the MPs who actively participated in the parliament of that year~(around 300 MPs depending on which year). We then extracted all bills sponsored by each MP and the corresponding votes for each bill. Lastly, we filtered out the yes votes for each ballot of the bills and which MPs voted yes. The data mining code is also available in the code repository of LAD~\footnote{\url{https://github.com/shenyangHuang/LAD}}.

We also provide more information on the political environment from 2006 to 2019. During this time period the government party in power has changed. In 2006, the government was minority Conservative until 2015 when the Liberal party won and formed a majority parliament~\cite{macfarlane2019renewed}. Studies have shown that minority governments appear to be less productive in legislative activity as consensus is harder to obtain~\cite{conley2011legislative}. Cohesion in the House of Commons within a political party during voting sessions are often observed and dissent has been seen amongst MPs of the same party who are less influential~\cite{garner2005party}. While elected to the House of Commons, MPs can sponsor more than one bill which can also include bills that may have not passed in prior parliaments. Parliament sessions follows no regular pattern from one parliament to another.

\section{Ablation Study on Choice of Power} \label{app:power}

\begin{figure}[ht]
    \begin{center}
        \centerline{\includegraphics[width=0.4\columnwidth,trim={0in 0in 0 0 },clip]{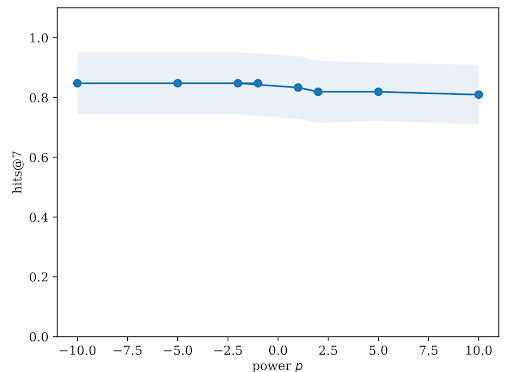}}\vspace*{-0.1in}
        \caption{The impact of power $p$ on the performance of \newmethod on the SBM experiment from Section~\ref{exp:limit} when $c_{in} = 12$ and $c_{out}=4$.}
     \label{Fig:power}
    \end{center}
\end{figure}

\changed{
Figure~\ref{app:power} shows the impact of the choice of power $p$ on the performance of \newmethod. We observe that \newmethod is largely robust to the choice of $p$ while the negative powers show slight performance improvement over the positive ones. Therefore, in this work, we choose the power $p=-10$.
}

\section{Additional Materials for \newmethod Experiments}
\changed{
This section contains additional material for \newmethod and multi-view experiments in Section~\ref{sec:multiview}. 
}
\begin{figure}[ht]
    \begin{center}
        \centerline{\includegraphics[width=0.8\columnwidth,trim={0in 0in 0 0 },clip]{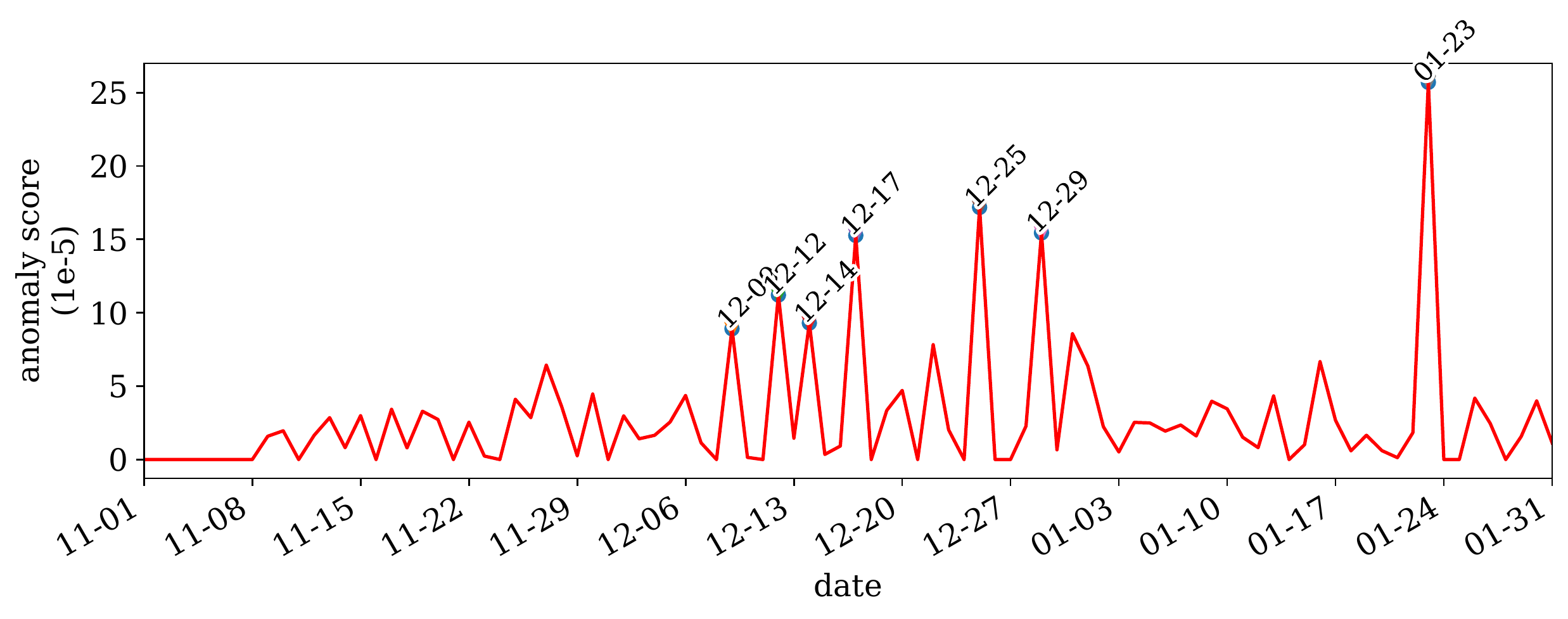}}\vspace*{-0.1in}
        \caption{NL maxLAD anomaly curve on NYC TLC 2015-2016 dataset  (annotated with the 7 most anomalous time points)}
     \label{Fig:max2016}
    \end{center}
\end{figure}

\begin{figure}[ht]
    \begin{center}
        \centerline{\includegraphics[width=0.8\columnwidth,trim={0in 0in 0 0 },clip]{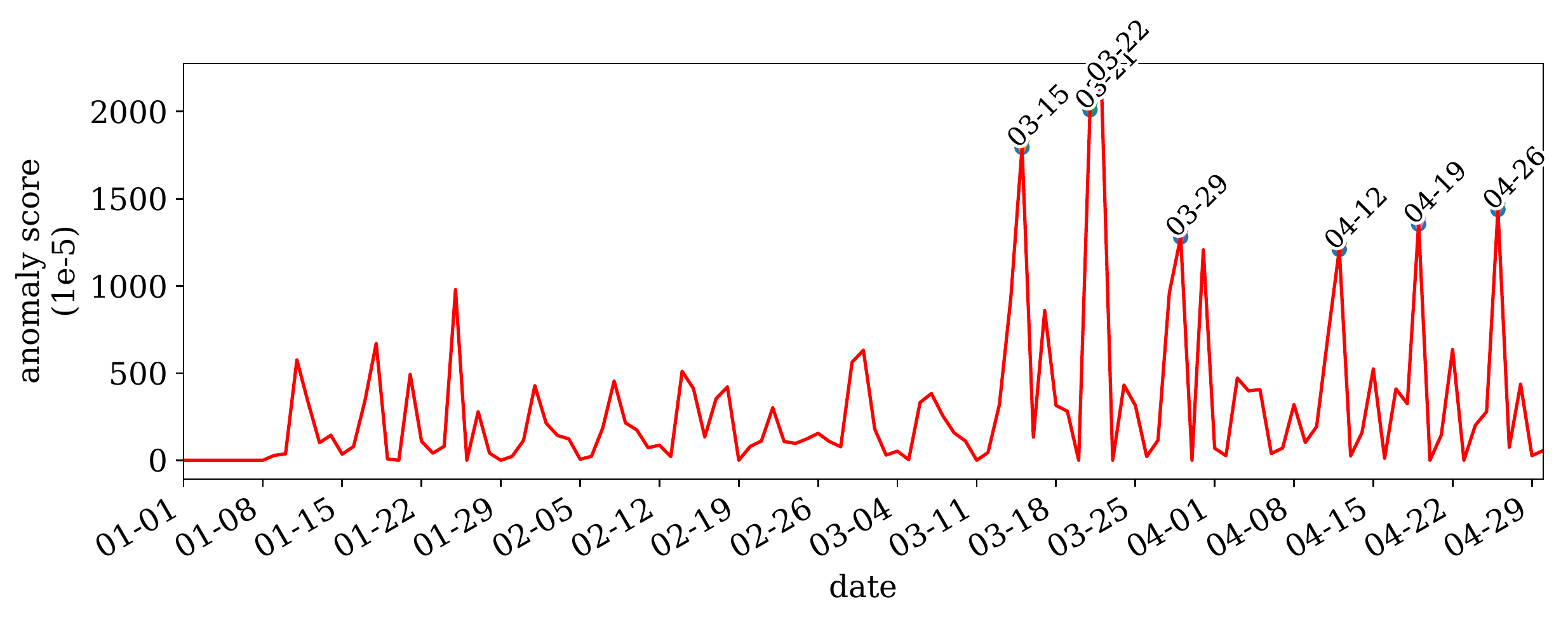}}\vspace*{-0.1in}
        \caption{NL maxLAD anomaly curve on NYC TLC 2020 dataset (annotated with the 7 most anomalous time points)}
     \label{Fig:max2020}
    \end{center}
\end{figure}

\begin{figure}[ht]
    \begin{center}
        \centerline{\includegraphics[width=0.8\columnwidth,trim={0in 0in 0 0 },clip]{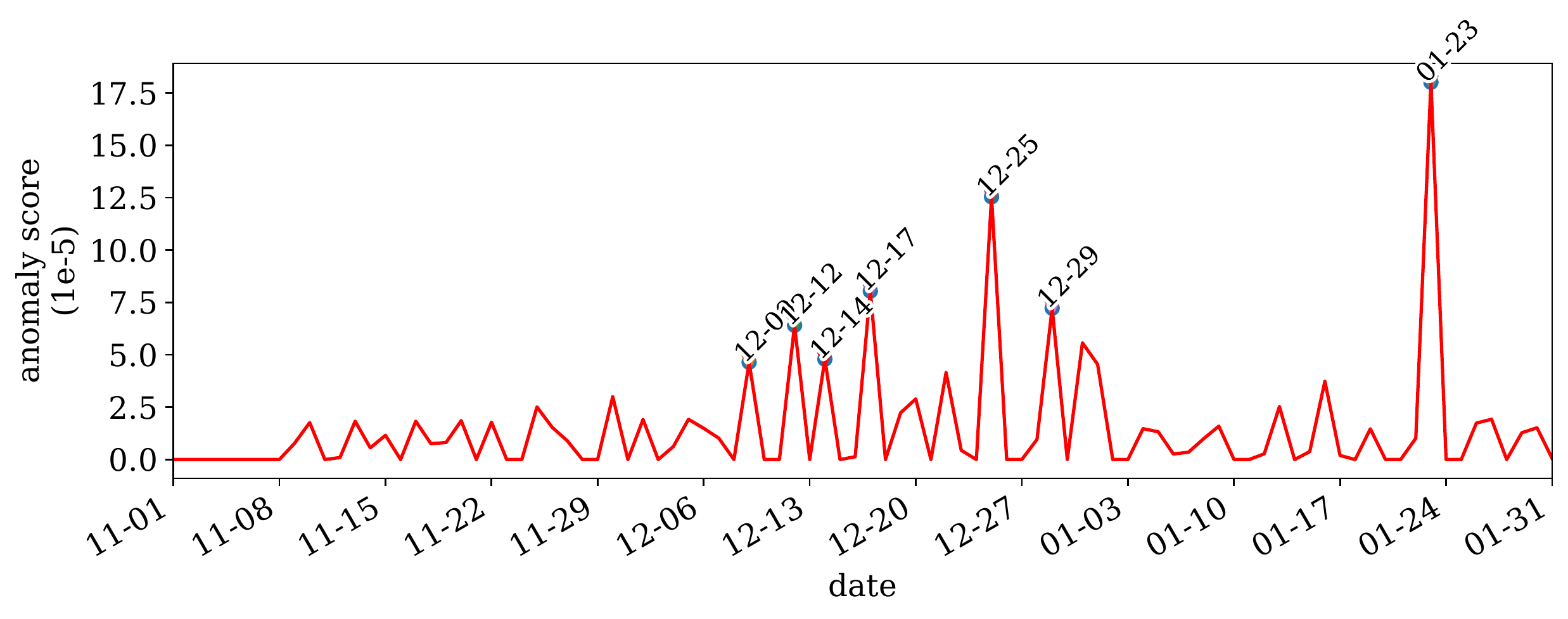}}\vspace*{-0.1in}
        \caption{NL meanLAD anomaly curve on NYC TLC 2015-2016 dataset (annotated with the 7 most anomalous time points)}
     \label{Fig:mean2016}
    \end{center}
\end{figure}

\begin{figure}[ht]
    \begin{center}
        \centerline{\includegraphics[width=0.8\columnwidth,trim={0in 0in 0 0 },clip]{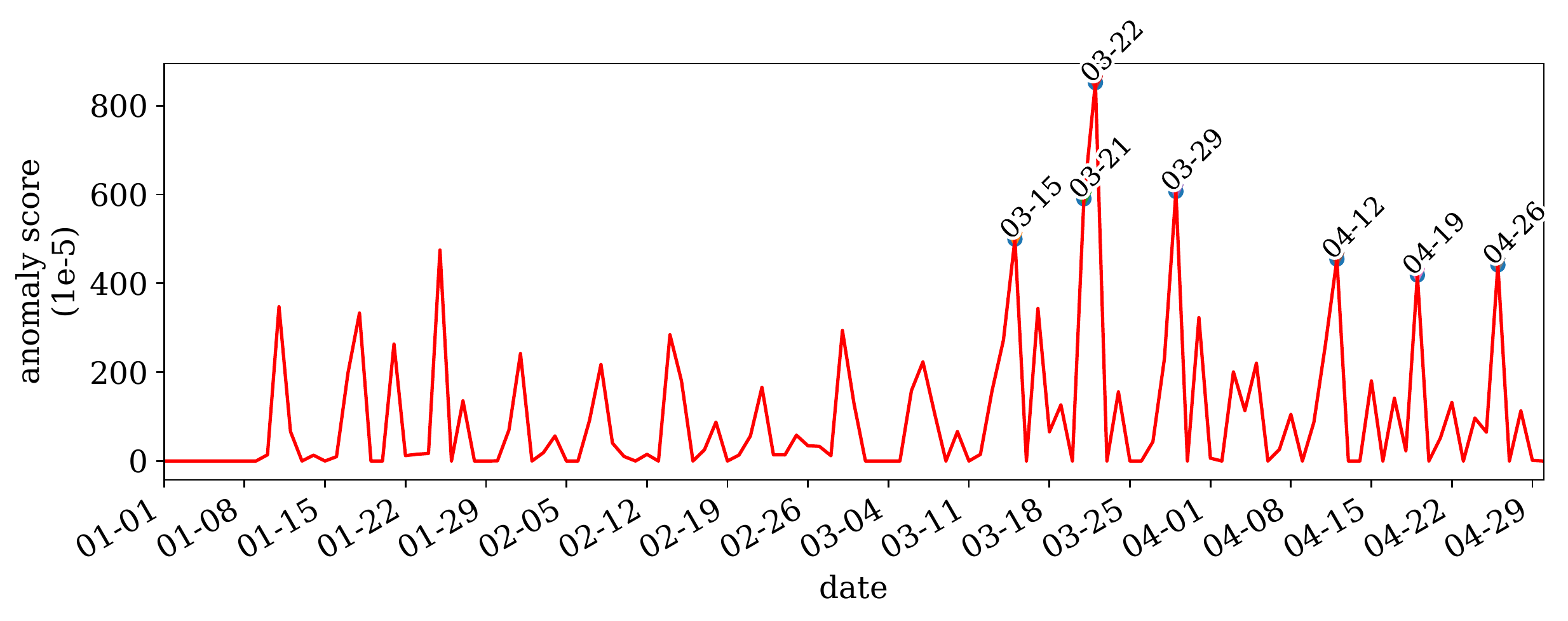}}    \vspace*{-0.1in}
        \caption{NL meanLAD anomaly score on NYC Transit 2020 dataset (annotated with the 7 most anomalous time points)}
     \label{Fig:mean2020}
    \end{center}
\end{figure}

\subsection{Additional Results}

\changed{
Figure~\ref{Fig:max2020} and~\ref{Fig:max2016}  shows the anomaly scores output by NL maxLAD for the NYC TLC 2020 and 2015-2016 case studies respectively. Figure~\ref{Fig:mean2016}, \ref{Fig:mean2020} shows the anomaly scores output by NL meanLAD on the same datasets.
}

\subsection{Data Processing} \label{app:dataprocessing}
\changed{
The cleaned and processed datasets are included with the submission. Gridlock days for 2016-2016 were obtained from The Official Website of the City of New York~\footnote{\url{https://www1.nyc.gov/events/gridlock-alert-day/13981/3}}. The dates related to NYC's COVID-19 response were obtained from a publicly available timeline\footnote{\href{https://www.investopedia.com/historical-timeline-of-covid-19-in-new-york-city-5071986}{https://www.investopedia.com/historical-timeline-of-covid-19-in-new-york-city-5071986}}.
}


\begin{table}[ht]
\centering 
\caption{ Key dates related to the 2015-2016 NYC TLC case study. Dates for gridlock days were sourced from the \href{https://www1.nyc.gov/events/gridlock-alert-day/13981/3}{official website} of the City of New York. Boldfaced dates correspond to those in \newmethod's 7 most anomalous days (labelled using \textsuperscript{†}) and those that are offset by one day (labelled with \textsuperscript{o}).}
\begin{minipage}[t]{0.4\columnwidth}
\begin{tabular}{ll}
20-Nov-15 & Gridlock Day \\
\rowcolor[rgb]{0.753,0.753,0.753} 25-Nov-15 & Gridlock Day \\
26-Nov-15 & Thanksgiving \\
\rowcolor[rgb]{0.753,0.753,0.753} 2-Dec-15 & Gridlock Day \\
4-Dec-15 & Gridlock Day \\
\rowcolor[rgb]{0.753,0.753,0.753} \textbf{11-Dec-15\textsuperscript{o}} & Gridlock Day \\
\textbf{16-Dec-15\textsuperscript{o}} & Gridlock Day \\
\end{tabular}
\end{minipage}%
\begin{minipage}[t]{0.4\columnwidth}
\begin{tabular}{ll}
\rowcolor[rgb]{0.753,0.753,0.753} \textbf{17-Dec-15\textsuperscript{†}} & Gridlock Day \\
\textbf{18-Dec-15\textsuperscript{o}} & Gridlock Day \\
\rowcolor[rgb]{0.753,0.753,0.753} 23-Dec-15 & Gridlock Day \\
\textbf{25-Dec-15\textsuperscript{†}} & Christmas Day \\
\rowcolor[rgb]{0.753,0.753,0.753} \textbf{31-Dec-15\textsuperscript{†}} & New Years Eve \\
\textbf{1-Jan-16\textsuperscript{o}} & New Years Day \\
\rowcolor[rgb]{0.753,0.753,0.753} \textbf{23-Jan-16\textsuperscript{†}} & Blizzard
\end{tabular}
\end{minipage}
\end{table}
\label{tab:2016Dates}

\begin{table}[ht]
\caption{ Key dates relating to the response of COVID-19 in New York State. Boldfaced dates correspond to those in \newmethod's 7 most anomalous days (labelled using \textsuperscript{†}) and those that are offset by one day (labelled with \textsuperscript{o}).}
\centering
\begin{tabular}{rp{0.8\columnwidth}}
7-Mar-20 & State of Emergency declared in New York State\\
\rowcolor[rgb]{0.753,0.753,0.753} 8-Mar-20 & Issued guidelines relating to public transit \\
12-Mar-20 & Events with more than 500 attendees are cancelled/postponed\\
\rowcolor[rgb]{0.753,0.753,0.753} \textbf{13-Mar-20\textsuperscript{o}} & National State of Emergency declared \\
16-Mar-20 & NYC public schools close \\
\rowcolor[rgb]{0.753,0.753,0.753} 17-Mar-20 & \cellcolor[rgb]{0.753,0.753,0.753}NYC bars and restaurants can only operate by delivery\\
\textbf{22-Mar-20\textsuperscript{†}} & ``NYS on Pause Program'' begins, all non-essential workers must stay home\\
\rowcolor[rgb]{0.753,0.753,0.753} \textbf{28-Mar-20\textsuperscript{o}} & All non-essential construction halted \\
\textbf{6-Apr-20\textsuperscript{o}} & Extension of of stay-at-home order and school closures\\
\rowcolor[rgb]{0.753,0.753,0.753} 16-Apr-20 & {\cellcolor[rgb]{0.753,0.753,0.753}}Extension of of stay-at-home order and school closures\\
30-Apr-20 & Subway ceases to operate during early hours
\end{tabular}
\label{tab:2020Dates}
\end{table}

\subsection{NYC TLC 2015-2016} \label{app:NYC2015}
\changed{
We used the longitude and latitude coordinates of pickup and drop-off locations to map each trip to the nearest pair of taxi zones. Doing so on a day-by-day basis generated a dynamic two-view directed weighted graphs over 264 nodes (taxi zones) in the greater NYC region, with each time point corresponding to a different day. Edge weights in these graphs correspond to the number of trips between two taxi zones without consideration for the number of passengers. We also point out that \citet{SPOTLIGHT} mapped longitude and latitude coordinates to a set of ``57 geographically or conceptually distinguishable zones based on common knowledge''~\cite{SPOTLIGHT}, which we were unable to replicate. They also chose to use hourly time points, though this does not drastically impact the comparison between \newmethod and SPOTLIGHT's results.
}

\changed{
\citet{SPOTLIGHT} also examined NYC TLC ridership data for November 2015 - January 2016. As mentioned in our Experiments section, there was a sizeable overlap in the dates identified by SPOTLIGHT and \newmethod. In terms of differences, \newmethod ranked two NYC gridlock days (dates that the Department of Transportation expected to have unusually busy traffic~(11\textsuperscript{th} and 17\textsuperscript{th} respectively, see Table~\ref{tab:2016Dates}) higher in its list of anomalous days. On the other hand, SPOTLIGHT ranked American Thanksgiving higher. 
}

\changed{
This might be partly explained by the fact that we use multi-view data while SPOTLIGHT focuses on single-view graphs. The differences in data pre-processing (see the ``NYC TLC 2015-2016'' subsection above) could also be partly responsible. We were also unable to ascertain whether they used both the green and yellow taxi ridership data or focused on one view only. We believe the combination of these factors and the large methodological differences between SPOTLIGHT and \newmethod go a long way in explaining the differences between the two sets of results.
}

\subsection{NYC TLC 2020} \label{app:NYC2020}
\changed{
The 2020 NYC TLC trip records already possessed taxi zone data for pickup and drop-off locations. We also note that two taxi zones had been added between 2015-2020, and that the daily directed weighted graphs spanned 266 nodes instead of 264. Key dates relating to the response of COVID-19 in New York State is detailed in Table~\ref{tab:2020Dates}.
}

\end{document}